\documentclass[sigconf, nonacm]{acmart}

\usepackage{booktabs} 

\usepackage{epstopdf}

\usepackage{setspace}
\usepackage{placeins}
\usepackage{afterpage}
\usepackage[utf8]{inputenc}
\usepackage{graphicx}
\usepackage{balance}
\usepackage{amsthm}
\usepackage{color}
\usepackage{caption}
\usepackage{amsmath}
\usepackage{tabu}
\usepackage{array}
\usepackage{booktabs}
\usepackage{threeparttable}
\usepackage{relsize}
\usepackage{physics}

\usepackage{algorithm}
\usepackage[noend]{algpseudocode}

\newlength{\maxwidth}
\newcommand{\algalign}[2]
{\makebox[\maxwidth][r]{$#1{}$}${}#2$}

\usepackage{float}
\usepackage{stfloats}
\usepackage{lipsum}
\usepackage[english]{babel}

\usepackage{comment}

\usepackage[T1]{fontenc}
\usepackage{mwe}    
\usepackage{subfig}

\let\oldnl\nl
\newcommand{\nonl}{\renewcommand{\nl}{\let\nl\oldnl}}
\newtheorem{thm}{Theorem}
\newtheorem{problem}{Problem}

\theoremstyle{definition}
\newtheorem{defn}{Definition}

\def\Equal{\texttt{=}}

\begin{document}
\title{Fair Spatial Indexing: A paradigm for Group Spatial Fairness}

\author{Sina Shaham}
\affiliation{%
  \institution{Viterbi School of Engineering\\University of Southern California}
  \state{Los Angeles, California} 
  \country{USA}
}
\email{sshaham@usc.edu}

\author{Gabriel Ghinita}
\affiliation{%
  \institution{College of Science and Engineering\\Hamad Bin Khalifa University\\Qatar Foundation, Doha, Qatar}
}
\email{gghinita@hbku.edu.qa}

\author{Cyrus Shahabi}
\affiliation{%
  \institution{Viterbi School of Engineering\\University of Southern California}
 \state{Los Angeles, California} 
  \country{USA}
}
\email{shahabi@usc.edu}




\begin{abstract}
Machine learning (ML) is playing an increasing role in decision-making tasks that directly affect individuals, e.g., loan approvals, or job applicant screening. Significant concerns arise that, without special provisions, individuals from under-privileged backgrounds may not get equitable access to services and opportunities. Existing research studies {\em fairness} with respect to protected attributes such as gender, race or income, but the impact of location data on fairness has been largely overlooked. With the widespread adoption of mobile apps, geospatial attributes are increasingly used in ML, and their potential to introduce unfair bias is significant, given their high correlation with protected attributes.
We propose techniques to mitigate location bias in machine learning. Specifically, we consider the issue of miscalibration when dealing with geospatial attributes. We focus on {\em spatial group fairness} and we propose a spatial indexing algorithm that accounts for fairness. Our KD-tree inspired approach significantly improves fairness while maintaining high learning accuracy, as shown by extensive experimental results on real data. 
\end{abstract}

\maketitle

\section{Introduction}



%

Recent advances in machine learning (ML) led to its adoption in numerous decision-making tasks that directly affect individuals, such as loan evaluation or job application screening. Several studies~\cite{pessach2022review,berk2021fairness,mhasawade2021machine} pointed out that ML techniques may introduce bias with respect to protected attributes such as race, gender, age or income. The last years witnessed the introduction of {\em fairness} models and techniques that aim to ensure all individuals are treated equitably, focusing especially on conventional protected attributes (like race or gender). However, the impact of geospatial attributes on fairness has not been extensively studied, even though location information is being increasingly used in decision-making for novel tasks, such as recommendations, advertising or ride-sharing. Conventional applications may also often rely on location data, e.g. allocation of local government resources, or crime prediction by law enforcement using geographical features. For example, the Chicago Police Department releases monthly crime datasets~\cite{chicago} and classifies neighborhoods based on their crime risk level. Subsequently, the risk level is used to determine vehicle and house insurance premiums, which are increased to reflect the risk level, and in turn, result in additional financial hardship for individuals from under-privileged groups.

Fairness for geospatial data is a challenging problem, due to two main factors: (i) data are more complex than conventional protected attributes such as gender or race, which are categorical and have only a few possible values; and (ii) the correlation between locations and protected attributes may be difficult to capture accurately, thus leading to hard-to-detect biases.


We consider the case of {\em group fairness}~\cite{dwork2018individual}, which ensures no significant difference in outcomes occurs across  distinct population groups (e.g., females vs. males). In our setting, groups are defined with respect to geospatial regions. The data domain is partitioned into disjoint regions, and each of them represents a group. All individuals whose locations belong to a certain region are assigned to the corresponding group. In practice, a spatial group can correspond to a zip code, a neighborhood, or a set of city blocks. Our objective is to support arbitrary geospatial partitioning algorithms, which can handle the needs of applications that require different levels of granularity in terms of location reporting. {\em Spatial indexing}~\cite{wang2020geological,eldawy2015spatialhadoop,zhang2016inverted} is a common approach used for partitioning, and numerous techniques have been proposed that partition the data domain according to varying criteria, such as area, perimeter, data point count, etc. We build upon  existing spatial indexing techniques, and adapt the partition criteria to account for the specific goals of fairness. By carefully combining geospatial and fairness criteria in the partitioning strategies, one can obtain spatial fairness while still preserving the useful spatial properties of indexing structures (e.g., fine-level clustering of the data).

Specifically, we consider a set of partitioning criteria that combines physical proximity and {\em calibration error}. Calibration is an essential concept in classification tasks which quantifies the quality of a classifier. Consider a binary classification task, such as a loan approval process. Calibration measures the difference between the observed and predicted probabilities of any given point being labeled in the positive class. If one partitions the data according to some protected attribute, then the expectation would be that the probability should be the same across both groups (e.g., males and females should have an equal chance, on aggregate, to be approved for a loan). If the expected and actual probabilities are different, that represents a good indication of unfair treatment.

Our proposed approach builds a hierarchical spatial index structure by using a composite split metric, consisting of both geospatial criteria (e.g., compact area) and miscalibration error. In doing so, it allows ML applications to benefit from granular geospatial information, while at the same time ensuring that no significant bias is present in the learning process.

%

\begin{figure*}[t]
	\subfloat[Neighborhoods Partitioning]{%
	\label{Fig: Architecture - a}
	\includegraphics[scale=.5]{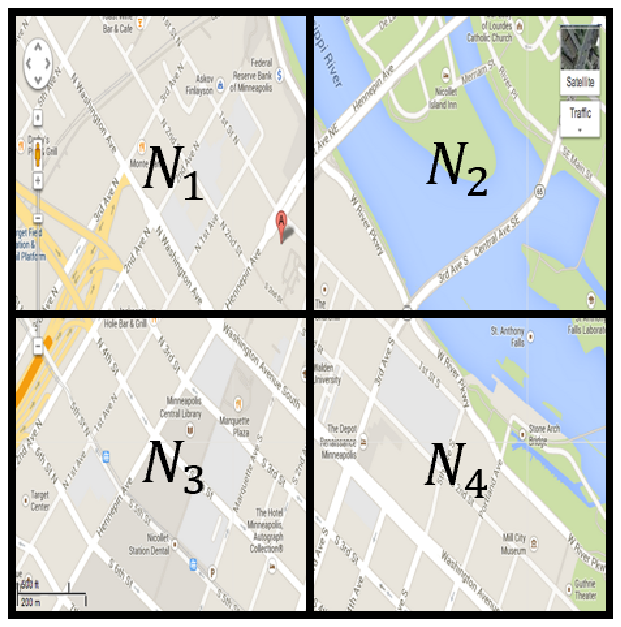}
	}
	\hfill
	\subfloat[Generation of Classifier Scores]{%
	\label{Fig: Architecture - b}
	\includegraphics[scale=.22]{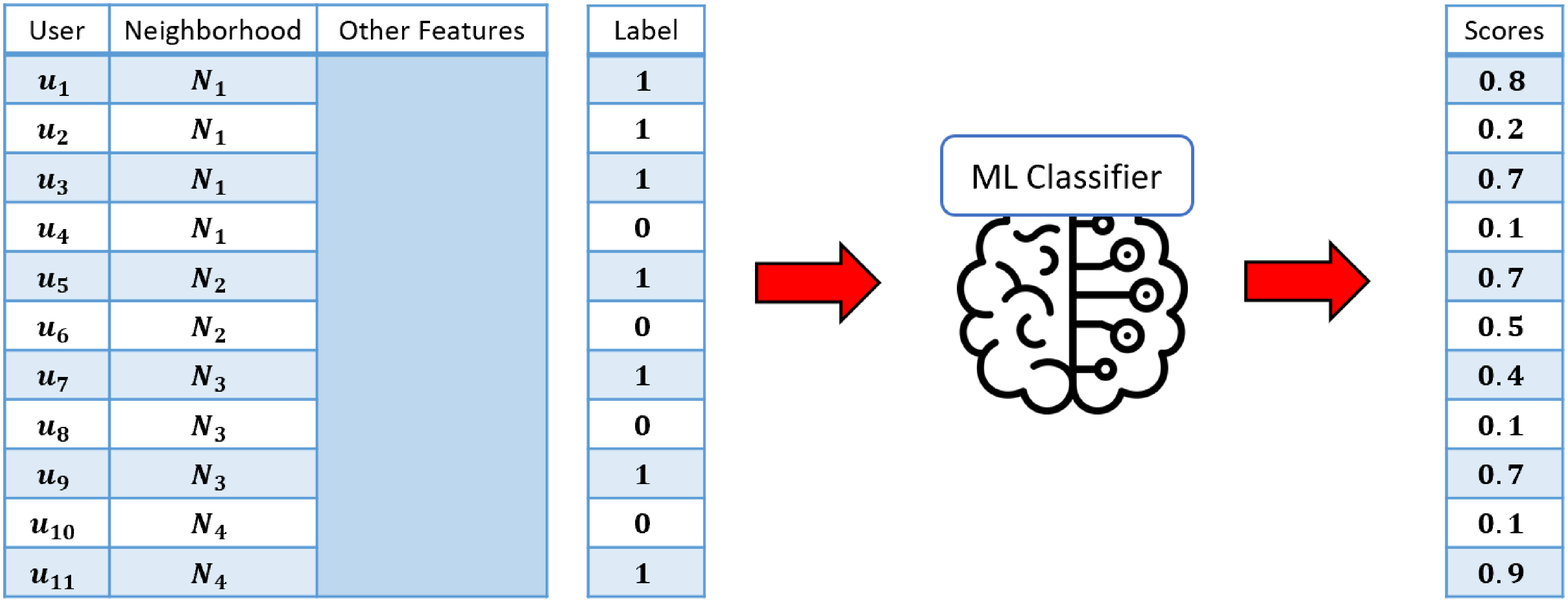}
	}
	\hfill
	\subfloat[Calibration of Neighborhoods]{%
	\label{Fig: Architecture - c}
	\includegraphics[scale=.4]{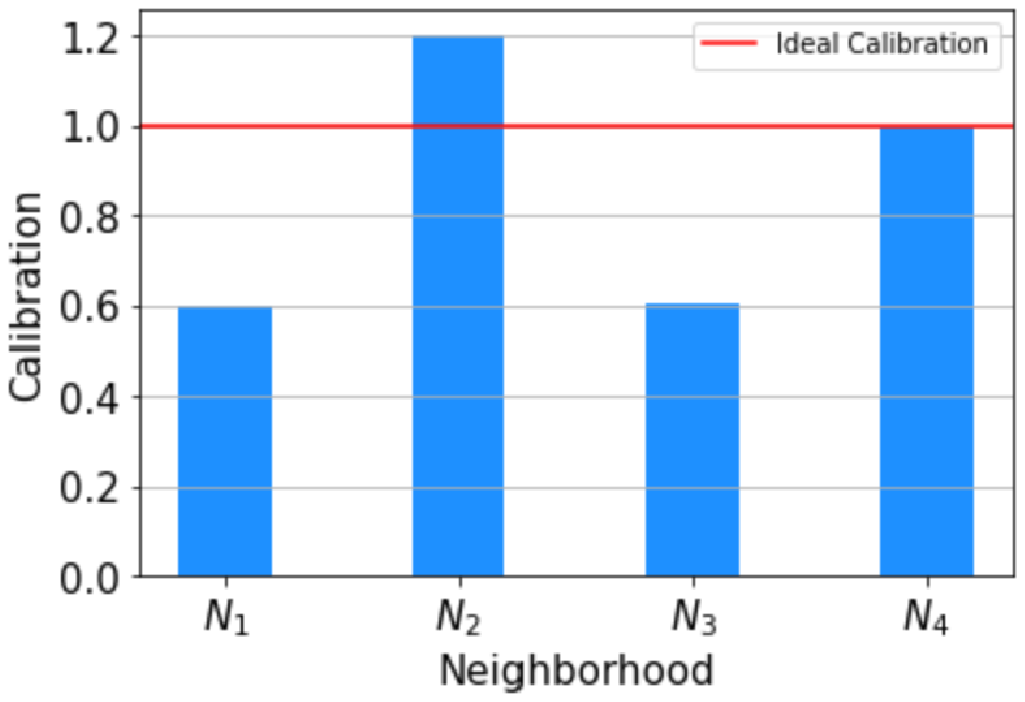}
	}
	\vspace{-10pt}
	\caption{An example of the miscalibration problem with respect to neighborhoods.}
	\label{Fig: Architecture}
	\vspace{-10pt}
\end{figure*}

Our specific contributions include:

\begin{itemize}
    \item We identify and formulate the problem of spatial group fairness, an important concept which ensures that geospatial information can be used reliably in a classification task, without introducing, intentionally or not, biases against individuals from underprivileged groups;
    
    \item We propose a new metric to quantify unfairness with respect to geospatial boundaries, called Expected Neighborhood Calibration Error (ENCE);
    
    \item We propose a technique for fair spatial indexing that builds on KD-trees and considers both geospatial and fairness criteria, by lowering miscalibration and minimizing ENCE;

    \item We perform an extensive experimental evaluation on real datasets, showing that the proposed approach is effective in enforcing spatial group fairness while maintaining data utility for classification tasks.

\end{itemize}

The rest of the paper is organized as follows: Section 2 provides background and fundamental definitions. Section 3 reviews related work. We introduce the proposed fair index construction technique in Section 4. Section 5 presents the results of our empirical evaluation, followed by conclusions in Section 6.

\section{Background}\label{Sec: System Model}

\subsection{System Architecture}

We consider a binary classification task $T$ over a dataset $D$ of individuals $u_1,\, ...,\, u_{|D|}$. The feature set recorded for $u_i$ is denoted by $\boldsymbol{x_i}\in \mathbb{R}^l $, and its corresponding label by $y_i\in \{0,1 \}$. Each record consists of $l$ features, including an attribute called {\em neighborhood}, which captures an individual's location, and is the main focus of our approach. The sets of all input data and labels are denoted by $\mathcal{X}$ and $\mathcal{Y}$, respectively. A classifier $h(.)$ is trained over the input data resulting in $h(\mathcal{X}) = (\hat{Y},\hat{S})$ where $\mathcal{\hat{Y}} = \{ \hat{y}_1,...,\hat{y}_{|D|}\}$ is the set of predicted labels ($\hat{y}_i\in \{0,1 \}$) and $\mathcal{S} = \{ s_1,...,s_{|D|}\}$ is the set of confidence scores ($s_i\in [0,1]$) for each label.  


The dataset's neighborhood feature indicates the individual's {\em spatial group}. We assume the spatial data domain is split into a set of partitions of arbitrary granularity. Without loss of generality, we consider a $U\times V$ grid overlaid on the map. The grid is selected such that its resolution captures adequate spatial accuracy as required by application needs. A set of neighborhoods is a  non-overlapping partitioning of the map that covers the entire space, with the $i^{th}$ neighborhood denoted by $N_i$, and the set of neighborhoods denoted by $\mathcal{N}$.

Figure~\ref{Fig: Architecture} illustrates the system overview. Figure~\ref{Fig: Architecture - a} shows the map divided into $4$ non-overlapping partitions $\mathcal{N} = \{ N_1,N_2,N_3,N_4 \}$. The neighborhood is recorded for each individual $u_1,...,u_{11}$ together with other features, and a classifier is trained over the data. The classifier's output is the confidence score for each entry which turns into a class label by setting a threshold.



\subsection{Fairness Metric}

Our primary focus is to achieve {\em spatial group fairness} using as metric the concept of {\em calibration}~\cite{naeini2015obtaining,pleiss2017fairness}, described in the following.

In classification tasks, it is desirable to have scores indicating the probability that a test data record belongs to a certain class. Probability scores are especially important in ranking problems, where top candidates are selected based on relative quantitative performance. Unfortunately, it is not granted that confidence scores generated by a classifier can be interpreted as probabilities. Consider a binary classifier that indicates an individual's chance of committing a crime after their release from jail (recidivism). If two individuals $u_1$ and $u_2$ get confidence scores $0.4$ and $0.8$, this cannot be directly interpreted as the likelihood of committing a crime by $u_2$ being twice as high as for $u_1$. The model {\em calibration} aims to alleviate precisely this shortcoming.

\begin{defn} \label{def: statistical parity}
(Calibration).  An ML model is said to be calibrated if it produces calibrated confidence scores. Formally, outcome score $R$ is {\em calibrated} if for all scores $r$ in support of $R$ it holds that
\begin{equation}
    P( y = 1 | R = r  )= r
\end{equation}
\end{defn}

This condition means that the set of all instances assigned a score value $r$ contains an $r$ fraction of positive instances. The metric is a group-level metric. Suppose there exist $10$ people who have been assigned a confidence score of $0.7$. In a well-calibrated model, we expect to have $7$ individuals with positive labels among them. Thus, the probability of the whole group is $0.7$ to be positive, but it does not indicate that every individual in the group has this exact chance of receiving a positive label.

To measure the amount of miscalibration for the whole model or for an output interval, the ratio of two key factors need to be calculated: expected confidence scores and the expected value of true labels. Abiding by the convention in~\cite{naeini2015obtaining}, we use functions $o(.)$ and $e(.)$ to return the true fraction of positive instances and the expected value of confidence scores, respectively. For example, the calibration of the model in Figure~\ref{Fig: Architecture - b} is computed as:
\begin{equation}
    \dfrac{e(h)}{o(h)} = \dfrac{(\sum_{u\in D} \hat{p}_u )/|D|}{(\sum_{u\in D} y_u )/|D|} = \dfrac{5.2/11}{7/11}  \approx .742
\end{equation}
Perfect calibration is achieved when a specific ratio is equal to one. Ratios that are above or below one are considered miscalibration cases. Another way to measure the calibration error is by using the absolute value of the difference between two values, denoted by $|e(h)- o(h)|$, with the ideal value being zero. In this work, the second method is utilized, as it eliminates the division by zero problem that may arise from neighborhoods with low populations.

\subsection{Problem Formulation}
Even when a model is overall well-calibrated, it can still lead to unfair treatment of individuals from different neighborhoods. In order to achieve spatial group fairness, we must have a well-calibrated model with respect to {\em all} neighborhoods. The existence of calibration error in a neighborhood can result in classifier bias and lead to systematic unfairness against individuals from that neighborhood (in Section~\ref{Sec: Experimental Evaluation}, we support this claim with real data measurements). 


\begin{defn} \label{def: calibration wrt neighborhoods}
(Calibration for Neighborhoods). Given neighborhood set $\mathcal{N} = \{ N_1,...,N_t \}$, we say that the score $R$ is calibrated in neighborhood $N_i$ if for all the scores $r$ in support of $R$ it holds that 

\begin{equation}
    P (y=1|R=r,N=N_i) = r, \;\;\;\;\;\;   \forall i\in [1,t]
\end{equation}
\end{defn}

 The following equations can be used to measure the amount of miscalibration with respect to neighborhood $N_i$,
\begin{equation}\label{Equation: subgroup calibration}
    \dfrac{e(h|N= N_i)}{o(h|N=N_i)}\;\;\;   \text{ or } \;\;\; |e(h|N= N_i) - o(h|N=N_i)|
\end{equation}

Going back to the example in Figure~\ref{Fig: Architecture}d, the calibration amount for neighborhoods $N_1$ to $N_4$ is visualized on a plot. Neighborhood $N_4$ is well-calibrated, whereas the others suffer from miscalibration. 





\begin{figure}[t]
\includegraphics[scale=.25]{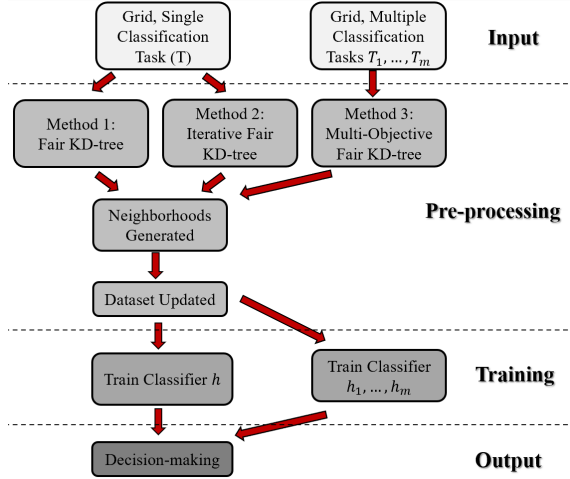}
\centering
\caption{Overview of the proposed mitigation techniques.}
\vspace{-15pt}
\label{Fig: Overview}
\end{figure}

\begin{problem}\label{problem formulation}
Given {\em m} binary classification tasks $T_1, T_2,...,T_m$, we seek to partition the space into continuous non-overlapping neighborhoods such that for each decision-making task, the trained model is well-calibrated for all neighborhoods. 
\end{problem}


\subsection{Evaluation Metrics}
A commonly used metric to evaluate the calibration of a model is Expected Calibration Error (ECE)~\cite{guo2017calibration}. The goal of ECE (detailed in Appendix~\ref{appendix: ecec}) is to  understand the validity of output confidence scores. However, our focus is on identifying the calibration error imposed on different neighborhoods. Therefore, we extend ECE and propose the Expected Neighborhood Calibration Error (ENCE) that captures the calibration performance over all neighborhoods.

\begin{defn} \label{def: statistical parity}
(Expected Neighborhood Calibration Error).  Given $t$ non-overlapping geospatial regions $\mathcal{N} = \{ N_1,...,N_t\}$ and a classifier $h$ trained over data located in these neighborhoods, the ENCE metric is calculated as:

\begin{align}
\text{ENCE} = \sum_{i=1}^t \dfrac{|N_i|}{|D|}| \text{o}(N_i) - e(N_i) |
\end{align}
where $o(N_i)$ and $e(N_i)$ return the true fraction of positive instances and the expected value of confidence scores for instances in $N_i$\footnote{Symbol $|.|$ denotes absolute value.}. 
\end{defn}


\newcommand{\rvec}{\mathrm {\mathbf {r}}} 
\begingroup
\begin{table}
\caption {Summary of Notations.} 
\centering
\begin{tabular}{>{\arraybackslash}m{2.7cm} >{\arraybackslash}m{5cm} }
\hline\hline
  Symbol  & Description\\    \hline
  $k$ & Number of features\\
   $D= \{ u_1,...,u_{|D|} \}$& Dataset of individuals\\
   $(x_i,y_i)$ & (Set of features, true label) for $u_i$\\
   $D = [\mathcal{X},\mathcal{Y}]$ & Dataset with user features and labels\\
   $ \mathcal{\hat{Y}} \Equal \{ \hat{y}_1,..,\hat{y}_{|D|}\}$& Set of predicted labels\\
    $\mathcal{S} = \{ s_1,...,s_{|D|}\}$ & Set of confidence scores\\
    $\mathcal{N} = \{ N_1, ..., N_t \}$ & Set of neighborhoods\\
    $U\times V$ & Base grid resolution\\
    $T$ & Binary classification task\\
    $m$ & Number of binary classification tasks\\
    $t$ & Number of neighborhoods\\
    $t_h$ & Tree height\\
\hline\hline
\end{tabular}
\label{tab:table1}
\end{table}
\endgroup

\section{Related Work}\label{Sec: Related Work}


\noindent{\bf Fairness Notions.}
There exist two broad categories of fairness notions~\cite{mehrabi2021survey,caton2020fairness}: individual fairness and group fairness. In group fairness, individuals are divided into groups according to a protected attribute, and a decision is said to be fair if it leads to a desired statistical measure across groups. Some prominent group fairness metrics are calibration~\cite{pleiss2017fairness},  statistical parity~\cite{kusner2017counterfactual}\cite{dwork2012fairness}, equalized odds~\cite{hardt2016equality}, treatment equality~\cite{berk2021fairness}, and test fairness~\cite{chouldechova2017fair}. Individual fairness notions focus on treating similar individuals the same way. Similarity may be defined with respect to a particular task~\cite{dwork2012fairness,joseph2016fairness}.

\noindent{\bf Spatial Fairness.}
Neighborhoods or individual locations are commonly used features for decision-making in government agencies, banks, etc. Unfairness may arise in  tasks such as mortgage lending~\cite{lee2021algorithmic}, job recruitment~\cite{faliagka2012integrated}, admission to schools~\cite{benabbou2019fairness}, and crime risk prediction~\cite{wang2022pursuit}. In~\cite{wang2022pursuit}, recidivism prediction models constructed using data from one location tend to perform poorly when they are used to predict recidivism in another location. 
The authors in~\cite{riederer2017price} formulate a loss function for individual fairness in social media and location-based advertisements. Pujol et al.~\cite{pujol2021equity} demonstrate the unequal impact of differential privacy on neighborhoods. Several attempts have been made to apply fairness notions for clustering datapoints in the Cartesian space. The notion in~\cite{kleindessner2020notion} defines clustering conducted for a point as fair if the average distance to the points in its own cluster is not greater than the average distance to the points in any other cluster. The authors in~\cite{mahabadi2020individual} focus on defining individual fairness for $k$-median and $k$-means algorithms. Clustering is defined to be individually fair if every point expects to have a cluster center within a particular radius.


\noindent{\bf Unfairness Mitigation} techniques can be categorized into three broad groups: pre-processing, in-processing, and post-processing. Pre-processing algorithms achieve fairness by focusing on the classifier's input data. Some well-known techniques include suppression of sensitive attributes, change of labels, reweighting, representation learning, and sampling~\cite{kamiran2012data}. In-processing techniques achieve fairness during training by adding new terms to the loss function~\cite{kamishima2012fairness} or including more constraints in the optimization. Post-processing techniques sacrifice the utility of output confidence scores and align them with the fairness objective~\cite{platt1999probabilistic}. 


\begin{figure}[t]
	\subfloat[Initial execution of classifier.]{%
	\includegraphics[scale=.25]{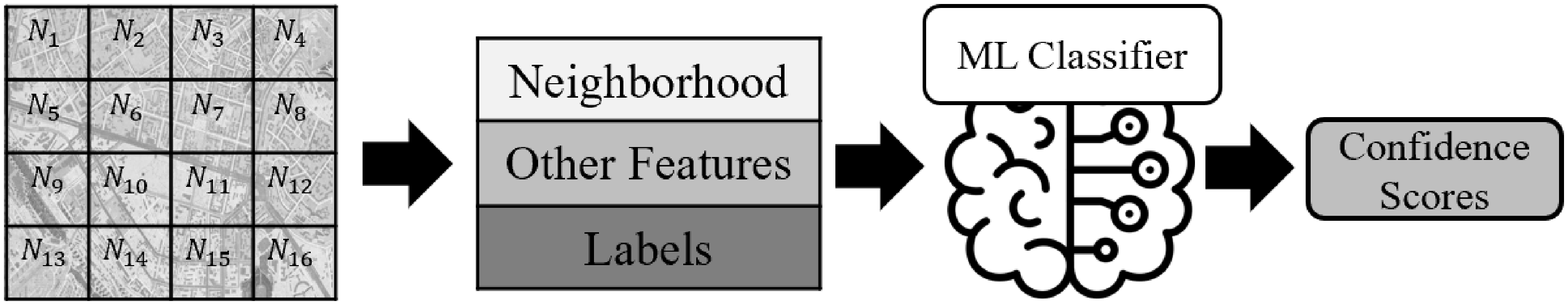}
	\label{Fig: kdtree bottom-up - a}
	}
	\hfill
	\subfloat[Re-districting the map based on fairness objective.]{%
	\includegraphics[scale=.3]{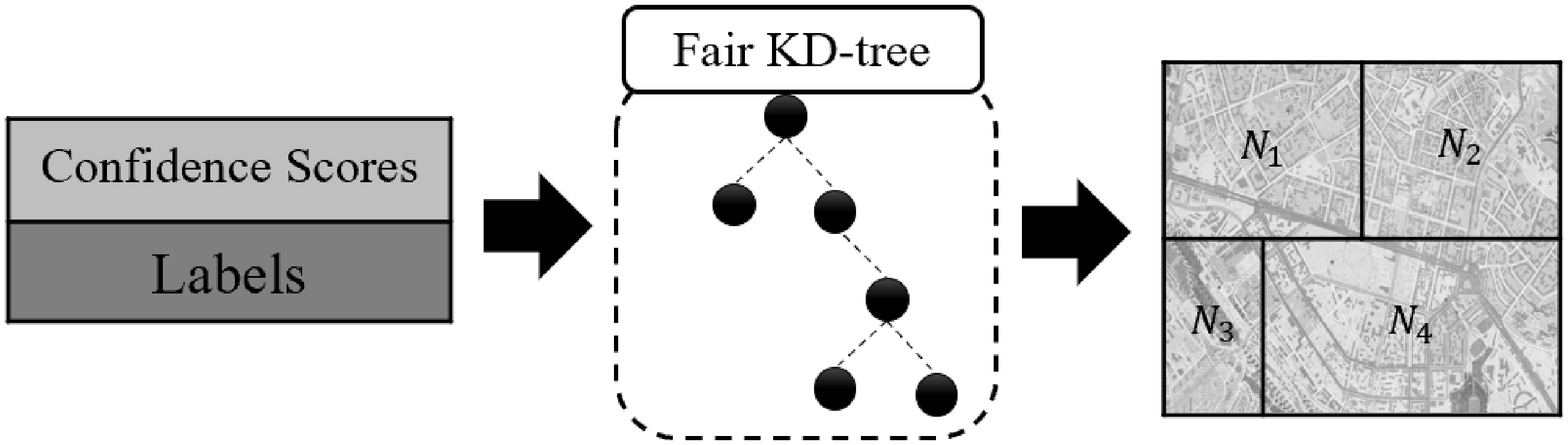}
	\label{Fig: kdtree bottom-up - b}
	}
	\hfill
	\subfloat[Training classifier based on re-districted neighborhoods.]{%
	\includegraphics[scale=.25]{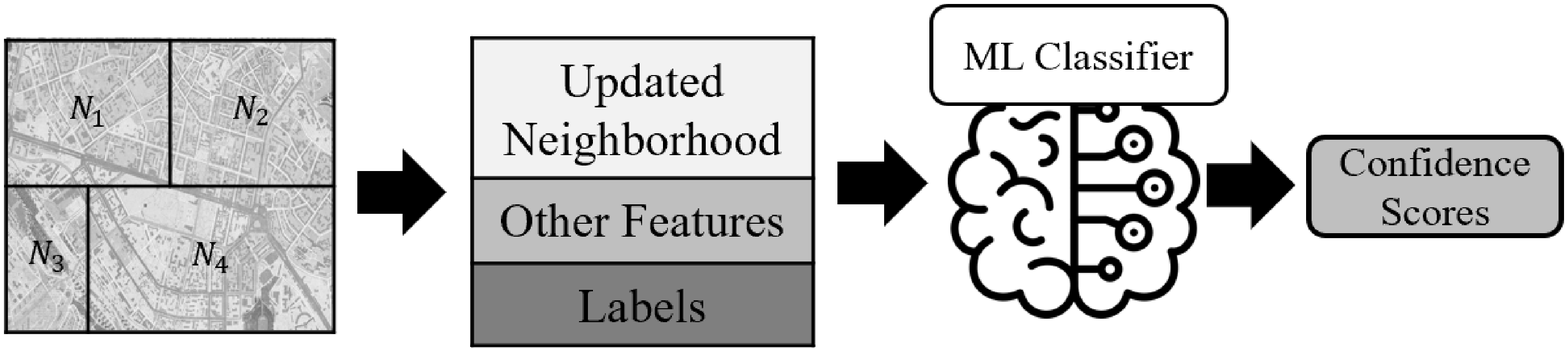}
	\label{Fig: kdtree bottom-up - c}
	}
	\vspace{-10pt}
	\caption{Overview of Fair KD-tree algorithm.}
	\vspace{-15pt}
	\label{Fig: Fair KD-tree}
\end{figure}

\section{Spatial Fairness through Indexing}


We introduce several algorithms that achieve group spatial fairness by constructing spatial index structures in a way that takes into account fairness considerations when performing data domain splits. We choose KD-trees as a starting point for our solutions, due to their ability to adapt to data density, and their property of covering the entire data domain (as opposed to structures like R-trees that may leave gaps within the domain).



Figure~\ref{Fig: Overview} provides an overview of the proposed solution.
Our input consists of a {\em base grid} with an arbitrarily-fine granularity overlaid on the map, the attributes/features of individuals in the data, and their classification labels. The attribute set includes individual location, represented as the grid cell enclosing the individual. 
We propose a suite of three alternative algorithms for fairness, which are applied in the pre-processing phase of the ML pipeline and lead to the generation of new neighborhood boundaries. Once spatial partitioning is completed, the location attribute of each individual is updated, and classification is performed again.


The proposed algorithms are:
\begin{itemize}
    \item {\em Fair KD-tree} is our primary algorithm and it re-districts spatial neighborhoods based on an initial classification of data over a base grid. Fair KD-tree can be applied to a single classification task.
    \item {\em Iterative Fair KD-tree} improves upon Fair KD-tree by refining the initial ML scores at every height of the index structure. It incurs higher computational complexity but provides improved fairness.
    \item {\em Multi-Objective Fair KD-tree} enables Fair KD-trees for multiple classification tasks. It leads to the generation of neighborhoods that fairly represent spatial groups for multiple objectives.
\end{itemize}



Next, we prove an important result that applies to all proposed algorithms, which states that any non-overlapping partitioning of the location domain has a weighted average calibration greater or equal to the overall model calibration. The proofs of all theorems are provided in Appendix~\ref{Sec: Appendix}. 


\begin{thm}\label{Theorem: calibration}
For a given model $h$ and a complete non-overlapping partitioning of the space $\mathcal{N} = \{ N_1, N_2,..., N_t \}$, ENCE is lower-bounded by the overall calibration of the model. 
\end{thm}

\noindent A broader statement can also be proven, showing that further partitioning leads to poorer ENCE performance. 

\begin{thm}\label{Theorem: calibration 2}
Consider a binary classifier $h$ and two complete non-overlapping partitioning of the space $\mathcal{N}_1$ and $\mathcal{N}_2$. If $\mathcal{N}_2$ is a sub-partitioning of $\mathcal{N}_1$, then:
\begin{equation}
\text{ENCE}(N_1) \leq \text{ENCE}(N_2)
\end{equation}
Neighborhood set $\mathcal{N}_2$ is a sub-partitioning of $\mathcal{N}_1$ if for every $N_i\in \mathcal{N}_1$, there exists a set of neighborhoods in $\mathcal{N}_2$ such that their union is $N_i$.
\end{thm}

\begin{algorithm}[t]
\caption{Fair KD-tree}\label{Algo: fair KD-tree}
\begin{algorithmic}[1]
\Statex {\color{black}  {\bf Input: } Grid ($U\times V$), Features ($\mathcal{X}$), Labels ($\mathcal{Y}$), Height ($t_h$).
\Statex { \bf Output:} New neighborhoods and updated feature set}
\Function{FairKDtree}{$N, \mathcal{X},\mathcal{Y},\mathcal{S}, t_h$}
    \If {$t_h=0$}
        \State $\mathcal{N} \leftarrow \mathcal{N} +N$
        \State \textbf{return} $True$
    \EndIf
    \State $axis\leftarrow \, t_h$  \textbf{mod} 2 
    \State $L_{k^*},\, R_{k^*}\leftarrow $ SplitNeighborhood$(N,\mathcal{Y},\mathcal{S}, axis) $
    \State Run FairKDtree($L_{k^*}, \mathcal{X}, \mathcal{Y}, \mathcal{S}, t_h-1$)   
    \State Run FairKDtree($R_{k^*}, \mathcal{X}, \mathcal{Y},\mathcal{S}, t_h-1$)
\EndFunction
\State $N_1 \leftarrow$ Grid
\State {\bf{Global}} $\mathcal{N}\leftarrow \{ \}$
\State Set all neighborhoods in $\mathcal{X}$ to $N_1$
\State Scores ($\mathcal{S}$) $\leftarrow$ Train ML model on $\mathcal{X}$ and $\mathcal{Y}$
\State Neighborhoods ($\mathcal{N})\leftarrow$ {\text Run} $FairKDtree(N_1, \mathcal{X},\mathcal{Y},\mathcal{S}, t_h) $
\State Update neighborhoods in $\mathcal{X}$
\State \textbf{return} $\mathcal{N},\, \mathcal{X}$
\end{algorithmic}
\end{algorithm}


\begin{algorithm}[t]
\caption{Split Neighborhood}\label{Algo: Split Neighborhood}
\begin{algorithmic}[1]
\Statex {\color{black}  {\bf Input: } Neighborhood ($N$), Confidence Scores ($\mathcal{S}$), Labels ($\mathcal{Y}$), Axis.
\Statex { \bf Output:} Non-overlapping split of $N$ into two neighborhoods}
\Function{SplitNeighborhood}{$N,\mathcal{S} ,\mathcal{Y},axis$}
    \If {$axis=1$}
    \State $N\leftarrow$ Transpose of $N$
    \EndIf
    \State $U'\times V'\leftarrow $ Dimensions of $N$
    \For {$k=1...U'$}
    \State $L_k \leftarrow $ Neighborhoods in $1...k$
    \State $R_k\leftarrow $ Neighborhoods in $k+1...U$
    \State $z_i \leftarrow$ Compute Equation (\ref{Equ: Fairness Objective}) for $L_k$ and $R_k$  \label{Algo line: obj call}
    \EndFor
    \State $k^*\leftarrow $ $\arg\min_{k}\;\; z_{k} $
    \State \textbf{return} $L_{k^*}$, $R_{k^*}$
\EndFunction
\end{algorithmic}
\end{algorithm}

\subsection{Fair KD-tree}

We build a KD-tree index that partitions the space into non-overlapping regions according to a split metric that takes into account the miscalibration metric within the regions resulting after each split. 
Figure~\ref{Fig: Fair KD-tree} illustrates this approach, which consists of three steps. Algorithm~\ref{Algo: fair KD-tree} presents the pseudocode of the approach.\\

{\em Step 1.} The base grid is used as input, where the location of each individual is represented by the identifier of their enclosing grid cell. This location attribute, alongside other features, is used as input to an ML classifier $h$ for training. The classifier's output is a set of confidence scores $\mathcal{S} $, as illustrated in Figure~\ref{Fig: kdtree bottom-up - a}. Once confidence scores are generated,  the true fraction of positive instances and the expected value of predicted confidence scores of the model with respect to neighborhoods can be calculated as follows: 
\begin{equation}
    e(h|N= N_i) = \dfrac{1}{|N_i|} (\sum_{u\in N_i} s_u )  \;\;\;\;\;\;   \forall i\in [1,t]
\end{equation}
\begin{equation}
    o(h|N= N_i) = \dfrac{1}{|N_i|} (\sum_{u\in N_i} y_u ) \;\;\;\;\;\;   \forall i\in [1,t]
\end{equation}
where $t$ is the number of neighborhoods.


\begin{algorithm}[t]
\caption{Iterative Fair KD-tree}\label{Algo: iterative fair KD-tree}
\begin{algorithmic}[1]
\Statex {\color{black}  {\bf Input: } Grid ($U\times V$), Features ($\mathcal{X}$), Labels ($\mathcal{Y}$), Height ($t_h$).
\Statex { \bf Output:} New neighborhoods and updated feature set}
\State $N_1 \leftarrow$ Grid
\State Set all neighborhoods in $\mathcal{X}$ to $N_1$
\State $\mathcal{N}\leftarrow \{N_1\}$
\While{$t_h>0$}
    \State Scores ($\mathcal{S}$) $\leftarrow$ Train ML model on $\mathcal{X}$ and $\mathcal{Y}$
    \State $\mathcal{N_{\text{new}}} \leftarrow \{ \empty \}$
    \For {$N_i\; \text{in}\; \mathcal{N}$}
        \State $L,R\leftarrow $ SplitNeighborhood$(N_i,\mathcal{S} ,\mathcal{Y},t_h\% 2) $
        \State $\mathcal{N_{\text{new}}} \leftarrow \mathcal{N_{\text{new}}} + {L,R}$
    \State 
    \EndFor $\mathcal{N} \leftarrow \mathcal{N_{\text{new}}}$
    \State Update neighborhoods in $\mathcal{X}$ based on $\mathcal{N}$
    \State $t_h\leftarrow t_h-1$
\EndWhile
\State \textbf{return} $\mathcal{N},\, \mathcal{X}$
\end{algorithmic}
\end{algorithm}

{\em Step 2.} This step performs the actual partitioning, by customizing the KD-tree split algorithm with a novel objective function. KD-trees are binary trees where a region is split into two parts, typically according to the median value of the coordinate across one of the dimensions (latitude or longitude). 
Instead, we select the division index that minimizes fairness metric, i.e., ENCE miscalibration. Confidence scores and labels resulted from the previous training step are used as input for the split point decision. For a given tree node, assume the corresponding partition covers $U'\times V'$ cells of the entire $U\times V$ grid. Without loss of generality, we consider partitioning on the horizontal axis (i.e., row-wise). The aim is to find an index $k$ which groups rows $1$ to $k$ into one node and rows $k+1$ to $U'$ into another, such that the  fairness objective is minimized.
Let $L_k$ and $R_k$ denote the left and right regions generated by splitting on index $k$. The fairness objective for index $k$ is:
\begin{equation}\label{Equ: Fairness Objective}
    z_k = \big{|}\; |L_k|\times |o(L_k)- e(L_k)| - |R_k|\times |o(R_k) - e(R_k)\; |\big{|} 
\end{equation}
In the above equation, $|L_k|$ and $|R_k|$ return the number of data entries in the left and right regions, respectively. The intuition behind the objective function is to minimize the model miscalibration difference as we heuristically move forward. Two key points about the above function are: (i) the formulation of calibration is used in linear format due to the possibility of a zero denominator, and (ii) the calibration values are weighted by their corresponding regions' cardinalities. The optimal index $k^*$ is selected as:
\begin{equation}
   k^* = \arg\min_{k}\;\; z_{k} \;\;\; 
\end{equation}

{\em Step 3.} On completion of the fair KD-tree algorithm, the index leaf set provides a non-overlapping partitioning of the map. In the algorithm's final step, the neighborhood of each individual in the dataset is updated according to the leaf set and used for training. 


The Fair KD-tree pseudocode is provided in Algorithms~\ref{Algo: fair KD-tree} and~\ref{Algo: Split Neighborhood}. The latter returns the split point based on the fairness objective, and it is being called several times in Algorithm~\ref{Algo: fair KD-tree}. This function will also be used within the Iterative KD-tree algorithm.

\begin{thm}\label{Theorem: cc fair kd-tree}
For a given dataset $D$, the required number of neighborhoods $t$ and the model $h$, the computational complexity of Fair KD-tree is $\mathcal{O}(|D|\times \lceil \log(t)\rceil)+ \mathcal{O}(h)$.
\end{thm}

\begin{figure*}[t]
\includegraphics[scale=.3]{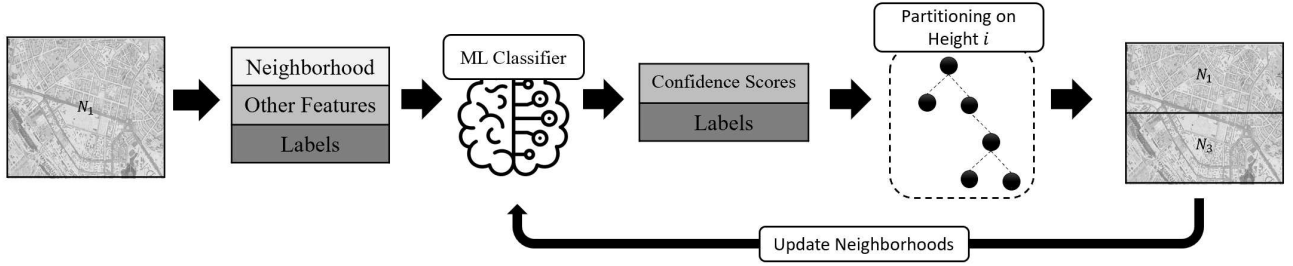}
\centering
\vspace{-10pt}
\caption{Overview of Iterative Fair KD-tree algorithm.}
\vspace{-15pt}
\label{Fig: new fitting example}
\end{figure*}

\begin{figure}[t]
\includegraphics[scale=.8]{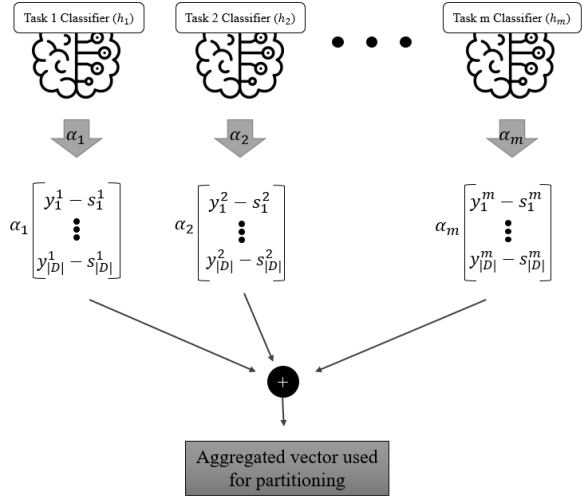}
\centering
\caption{Aggregation in Multi-Objective Fair KD-tree.}
\vspace{-15pt}
\label{Fig: multi}
\end{figure}


\subsection{Iterative Fair KD-tree}

One drawback of the Fair KD-tree algorithm is its sensitivity to the initial execution of the model, which uses the baseline grid to generate confidence scores. Even though the space is recursively partitioned following the initial steps, the scores are not re-computed until the index construction is finalized. The {\em iterative} fair KD-tree addresses this limitation by re-training the model and computing updated confidence scores after each split (i.e., at each level of the tree). A refined version of ML scores is used at every height of the tree, leading to a more fair redistricting of the map.

Similar to the Fair KD-tree algorithm, the baseline grid is initially used, and all grid cells are considered to be in the same neighborhood (i.e., a single spatial group covering the entire domain). The algorithm is implemented in $t_h$ iterations with the root node corresponding to the initial point (entire domain). As opposed to the Fair KD-tree algorithm that follows Depth First Search (DFS) recursion, the Iterative Fair KD-tree algorithm is based on Breadth First Search (BFS) traversal. Therefore, all nodes in the given height $i-1$ are completed before moving forward to the height $i$. Suppose we are in the $i^{th}$ level of the tree, and all nodes at that level are generated. Note that, the set of nodes at the same height represents a non-overlapping partitioning of the grid. The algorithm continues by updating the neighborhoods at height $i$ based on the $i-1$ level partitioning. Then, the updated dataset is used to train a new model, thus updating confidence scores for each individual. 

Algorithm~\ref{Algo: iterative fair KD-tree} presents the Iterative Fair KD-tree algorithm. Let $\mathcal{N}$ denote the set of all neighborhoods at level $i$ of the tree. For each neighborhood $N_i \in \mathcal{N}$, Iterative Fair KD-tree splits the region $N_i$ by calling the $SplitNeighborhood$ function in Algorithm~\ref{Algo: Split Neighborhood}. The split is done on the $x$-axis if $i$ is even and on the $y$-axis otherwise.

The algorithm provides a more effective way of determining a fair neighborhood partitioning by re-training the model at every tree level, but incurs higher computation complexity.

\begin{thm}\label{Theorem: cc iterative fair kd-tree}
For a given dataset $D$, the required number of neighborhoods $t$ and the model $h$, the computational complexity of Iterative Fair KD-tree is $\mathcal{O}(|D|\times \lceil \log(t)\rceil)+ \lceil \log(t)\rceil \times \mathcal{O}(h)$.
\end{thm}

\subsection{Multi-Objective Fair KD-tree}\label{section Multi-Objective Fair Partitioning}

So far, 
we focused on achieving a fair representation of space given a {\em single} classification task. In practice, applications may dictate multiple classification objectives. For example, a set of neighborhoods that are fairly represented in a city budget allocation task may not necessarily result in a fair representation of a map for deriving car insurance premia. 
Next, we show how Fair KD-tree can be extended to incorporate multi-objective decision-making tasks.


We devise an alternative method to compute initial scores in Line~\ref{Algo line: obj call} of Algorithm~\ref{Algo: Split Neighborhood}, which can then be called as part of Fair KD-tree in Algorithm~\ref{Algo: fair KD-tree}. A separate classifier is trained over each task to incorporate all classification objectives. Let $h_i$ be the $i^{th}$ classifier trained over $D$ and label set $\mathcal{Y}_i$ representing the task $T_i$. The output of the classifier is denoted by $\mathcal{S}_i = \{ s_1^i,...,s_{|D|}^i \}$, where in $s_j^i$, the superscript identifies the set $\mathcal{S}_i$ and the subscript indicates individual $u_j$. Once confidence scores for all models are generated, an auxiliary vector is constructed as follows:

\vspace{-10pt}
\begin{align}
\boldsymbol{v}_i &= \begin{bmatrix}
       s_1^i - y_1^i\\
       s_2^i - y_2^i\\
       \vdots \\
       s_{|D|}^i - y_{|D|}^i
     \end{bmatrix}, \;\;\; \; \; \; \; \forall i\in [1...t]
\end{align}

To facilitate task prioritization, hyper-parameters $\alpha_1, ...,\alpha_m$ are introduced such that $\sum_{i\Equal 1}^m \alpha_i= 1$ and $0 \leq \alpha_i \leq 1$. Coefficient $\alpha_i$ indicates the significance of classification $T_i$. The complete vector used for computing the partitioning is then calculated as,
\begin{equation}
    \boldsymbol{v}_{tot} = \sum_{i\Equal 1}^m \alpha_i \boldsymbol{v}_i = 
    \begin{bmatrix}
       \sum_{i\Equal 1}^m \alpha_i (s_1^i - y_1^i)\\
       \sum_{i\Equal 1}^m \alpha_i (s_2^i - y_2^i)\\
       \vdots \\
       \sum_{i\Equal 1}^m \alpha_i (s_{|D|}^i - y_{|D|}^i)
     \end{bmatrix}
\end{equation}

In the above formulation, each row corresponds to a unique individual and captures its role in all classification tasks. Let  $\boldsymbol{v}_{tot} [u_i] $ denote the entry corresponding to $u_i$ in $\boldsymbol{v}_{tot}$. Then the classification objective function in Eq.~\ref{Equ: Fairness Objective} is returned by:

\begin{equation}\label{Equ: Fairness Objective 3}
    z_k = \big{|}\; |L_k|\times  |
    \sum_{u_i\Equal L_k} \boldsymbol{v}_{tot} [u_i] |\; - \; |R_k|\times |\sum_{u_i\Equal R_k}   \boldsymbol{v}_{tot} [u_i] \; |\big{|} 
\end{equation}
and the optimal split point is selected as,
\begin{equation}
   k^* = \arg\min_{k}\;\; z_{k} \;\;\; 
\end{equation}

Vector aggregation is illustrated in Figure~\ref{Fig: multi}.

\begin{thm}\label{Theorem: cc multi objective fair kd-tree}
For a given dataset $D$, the required number of neighborhoods $t$ and $m$ classification tasks modelled by $h_1,...,h_m$, computational complexity of Multi-Objective Fair KD-tree is $\mathcal{O}(|D|\times \lceil \log(t)\rceil)+ \sum_{i\Equal1}^m\mathcal{O}(h_i)$.
\end{thm}

\section{Experimental Evaluation}\label{Sec: Experimental Evaluation}

\begin{figure*}[tbh]
	\hfill
	\subfloat[EdGap (Los Angeles)\label{figure: disparity proof s4}]{%
	\includegraphics[scale = 0.3]{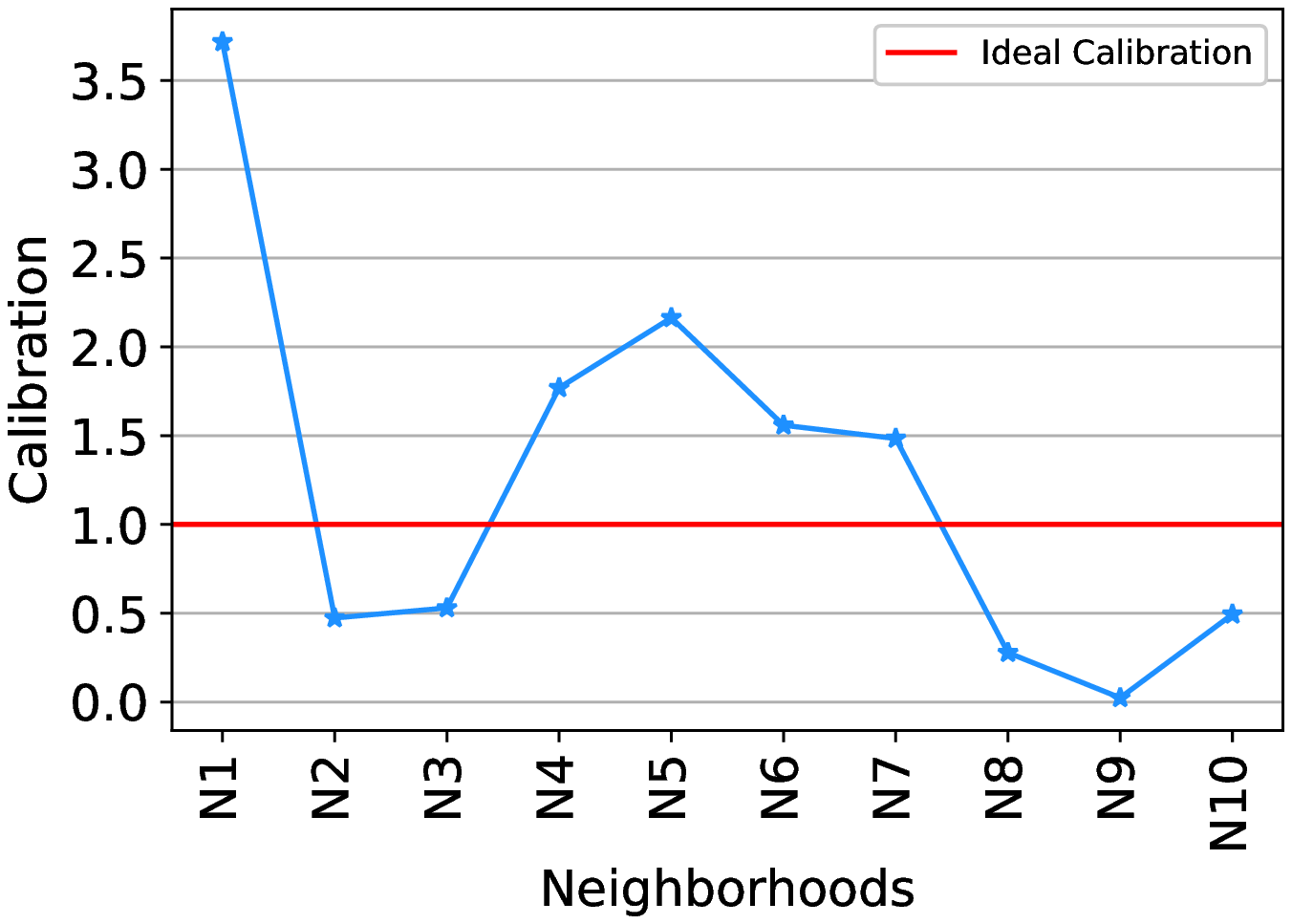}
	}
	\hfill
	\subfloat[EdGap (Los Angeles)\label{figure: disparity proof s5}]{%
	\includegraphics[scale = 0.3]{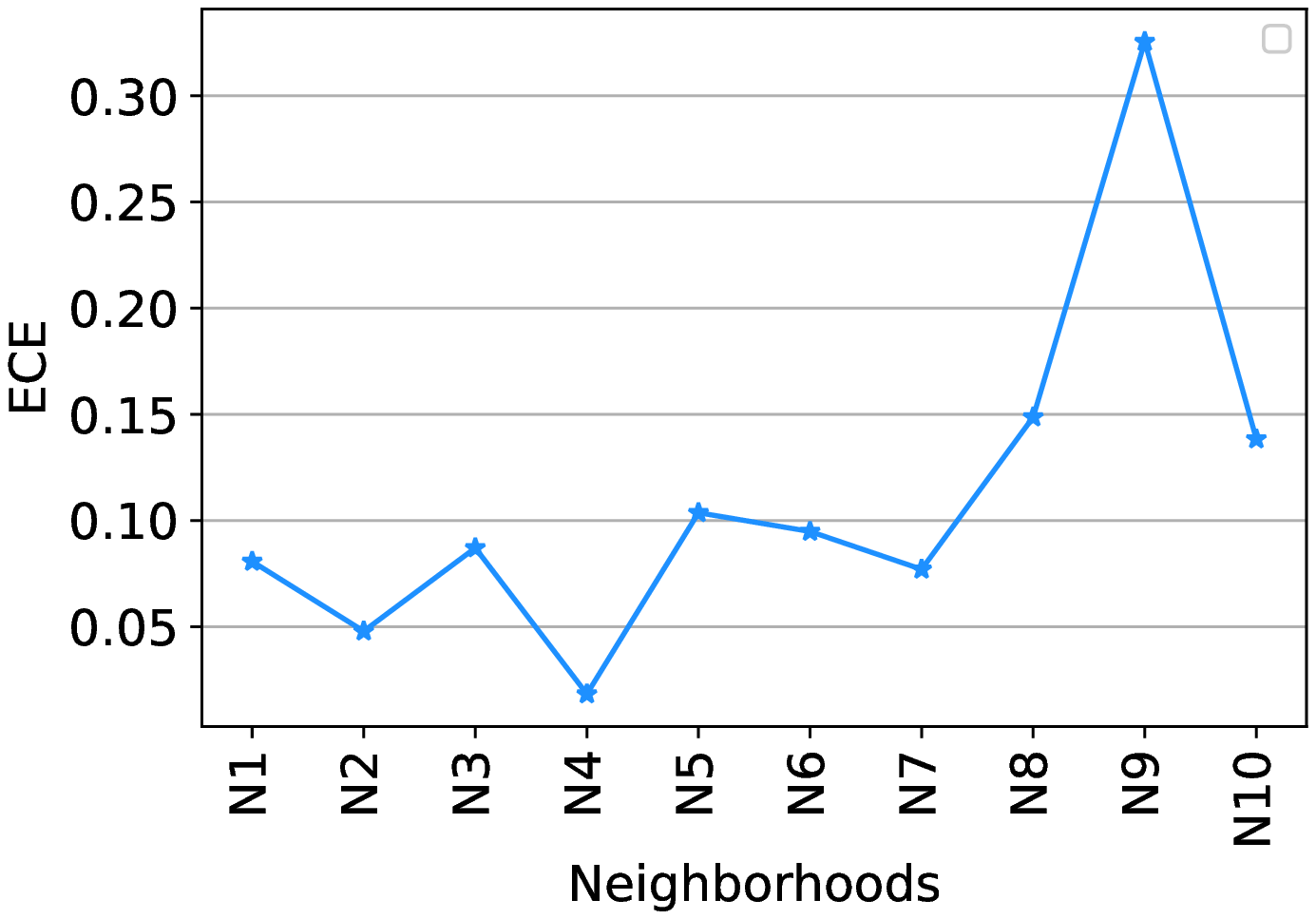}
	}
	\hfill
	\subfloat[EdGap (Houston)\label{figure: disparity proof s7}]{%
	\includegraphics[scale = 0.3]{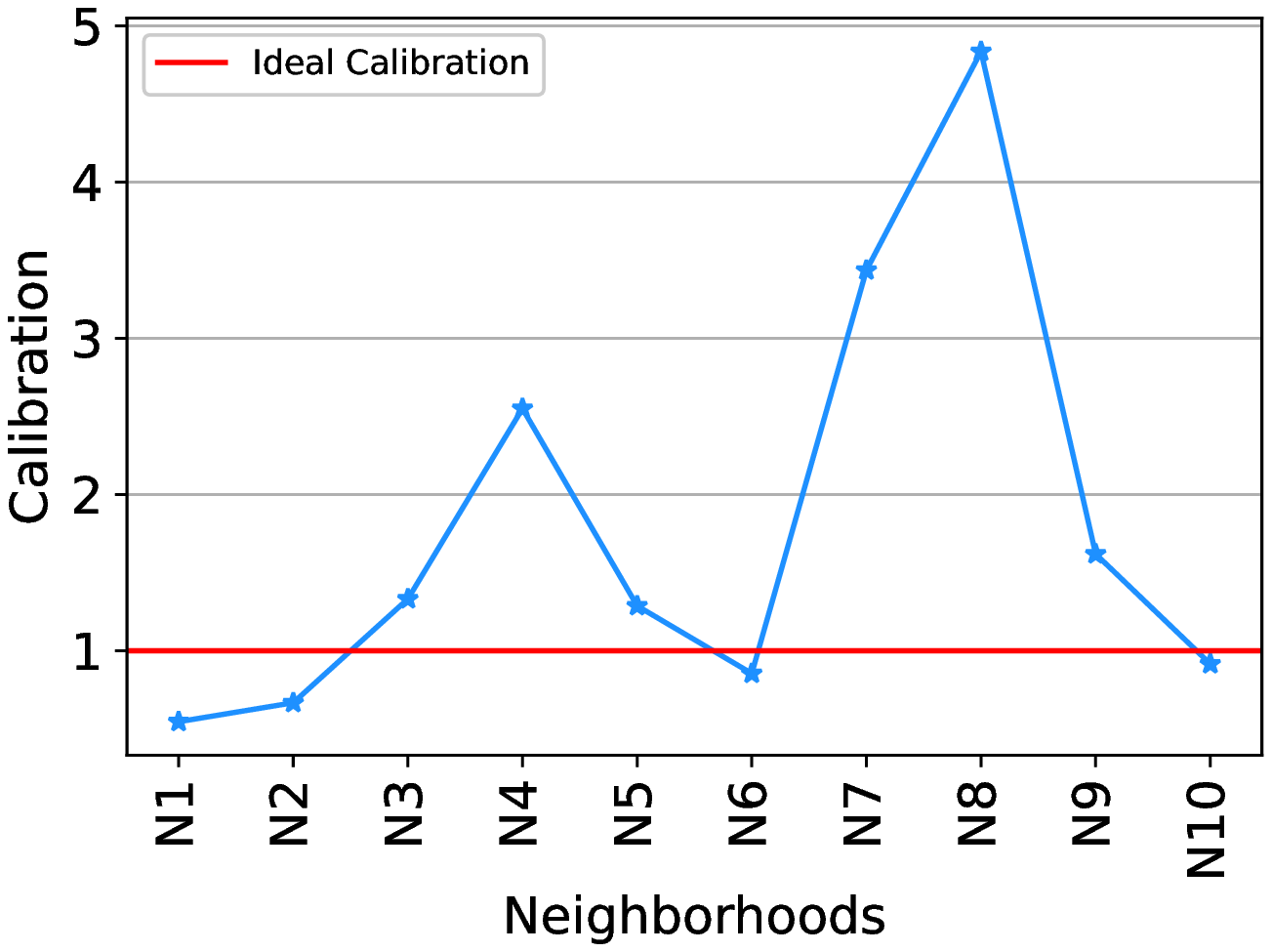}
	}
	\hfill
	\subfloat[EdGap (Houston)\label{figure: disparity proof s8}]{%
	\includegraphics[scale = 0.3]{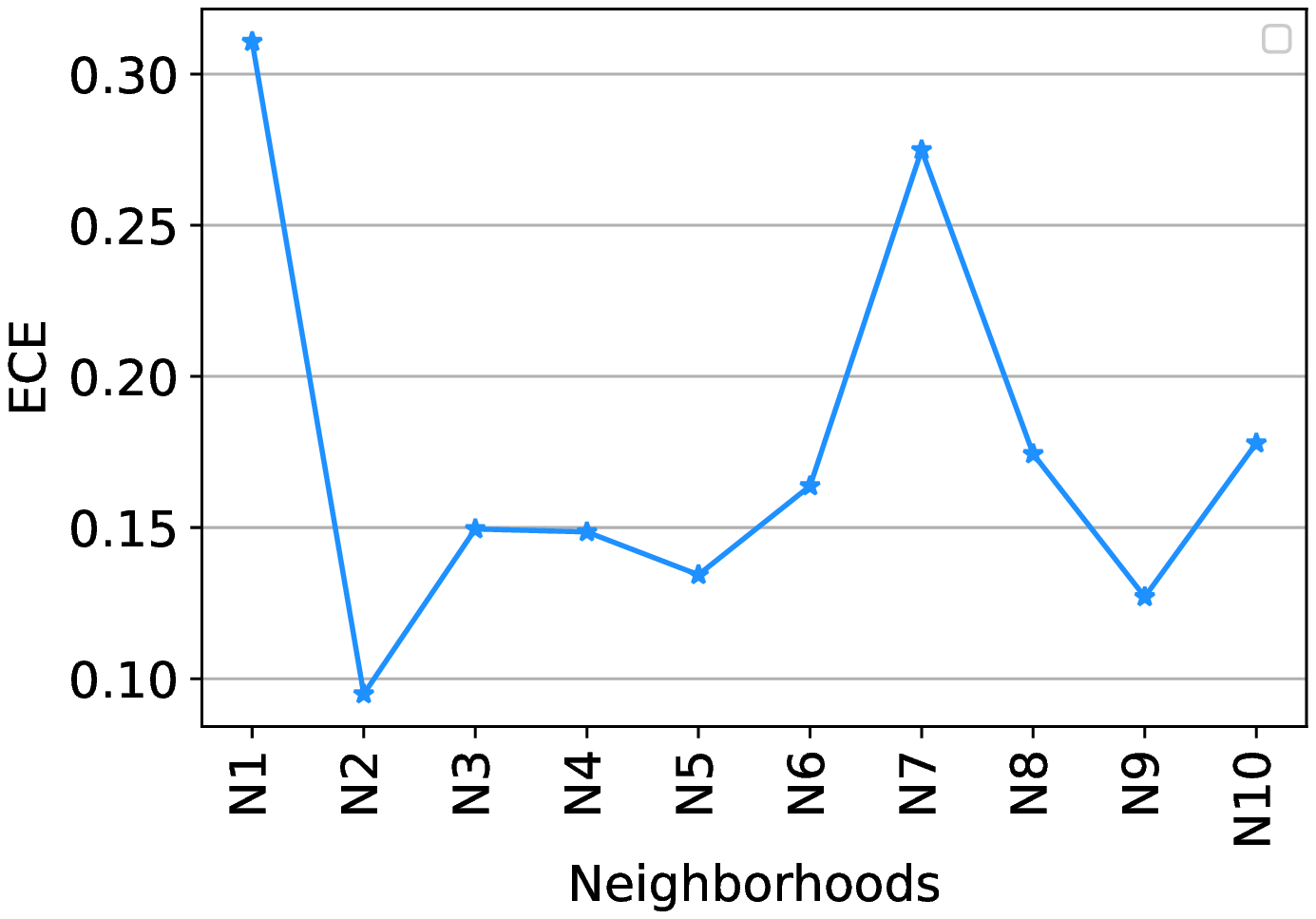}
	}
	\hfill
	\vspace{-10pt}
	\caption{Evidence of Model Disparity on Geospatial Neighborhoods.}
    \label{figure: disparity proof}
	\vspace{-10pt}
\end{figure*}

\begin{figure*}[tbh]
	\subfloat[Los Angeles (Logistic Regression)\label{figure: disparity proof s4}]{%
	\includegraphics[scale = 0.35]{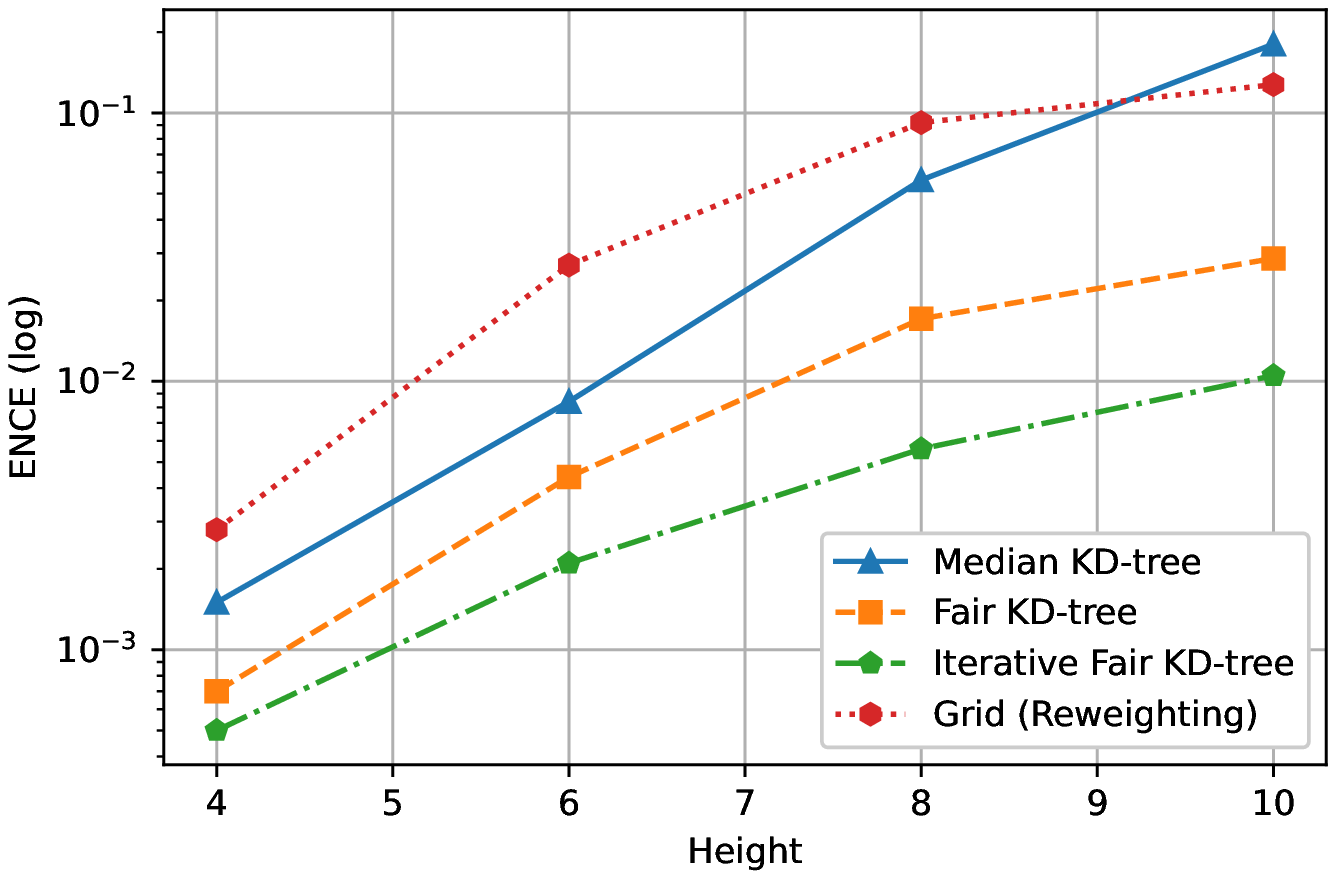}
	}
	\hfill
	\subfloat[Los Angeles (Decision Tree)\label{figure: disparity proof s5}]{%
	\includegraphics[scale = 0.35]{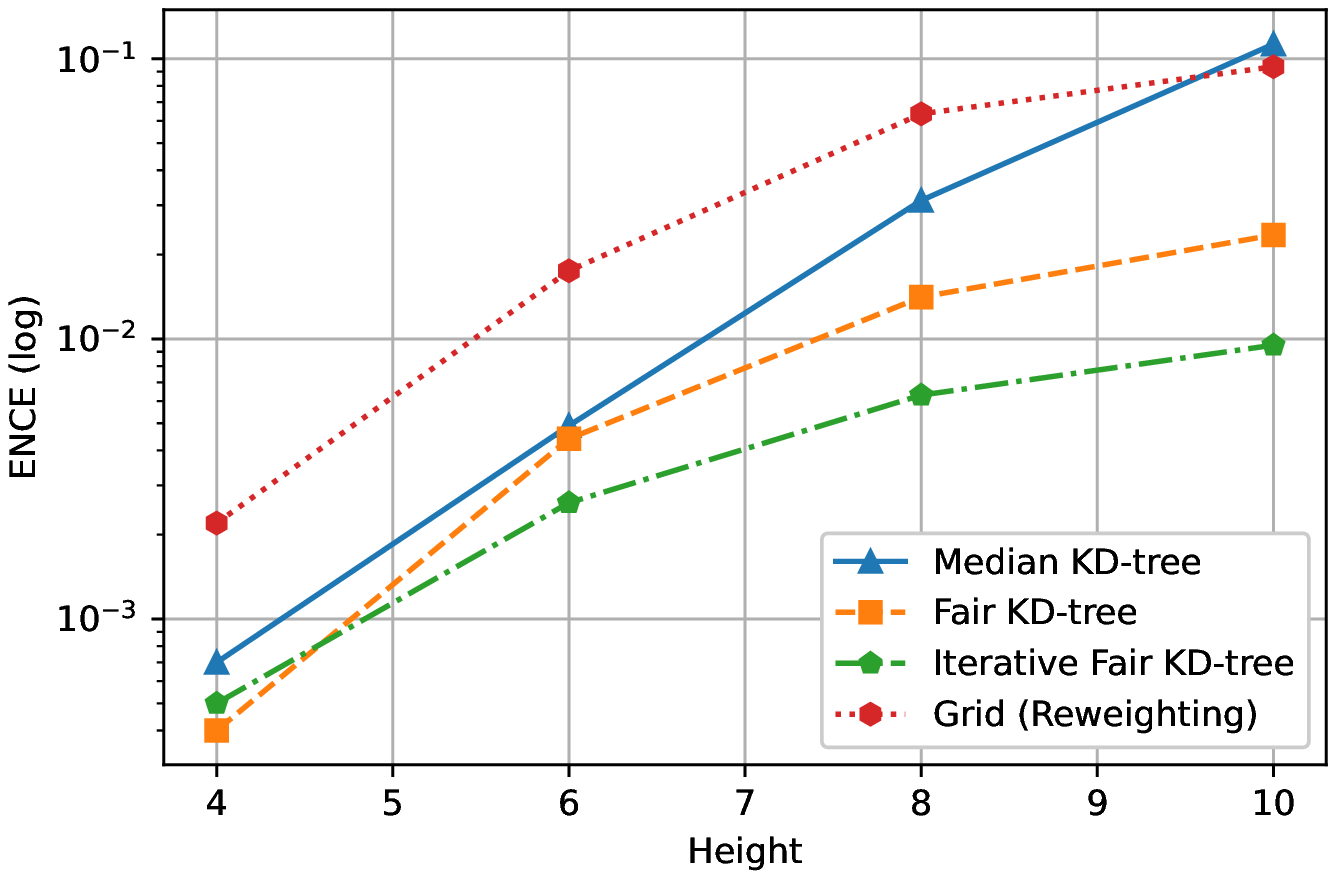}
	}
	\hfill
	\subfloat[Los Angeles (Naive Bayes)\label{figure: disparity proof s6}]{%
	\includegraphics[scale = 0.35]{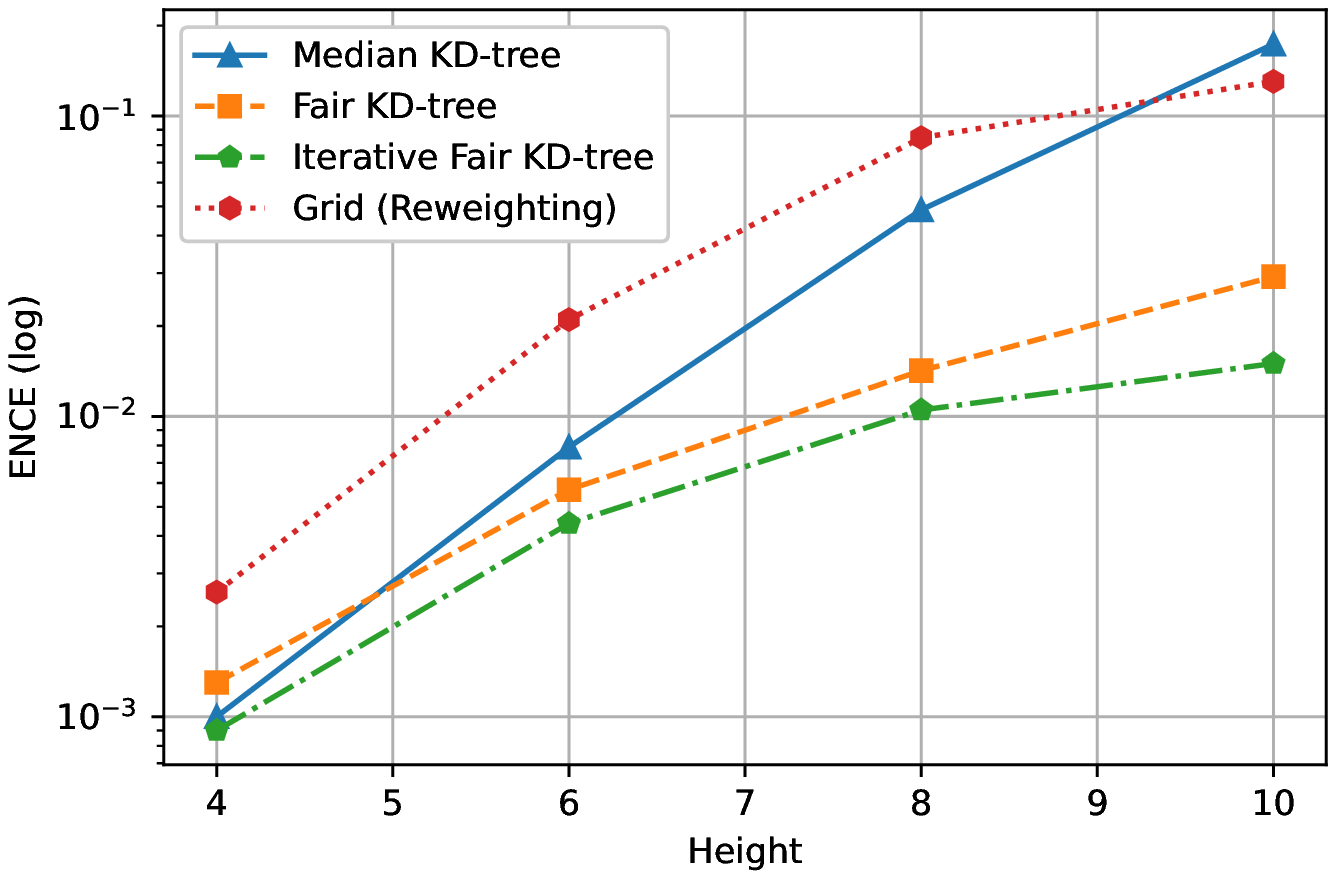}
	}
	\hfill
	\subfloat[Houston (Logistic Regression)\label{figure: disparity proof s7}]{%
	\includegraphics[scale = 0.35]{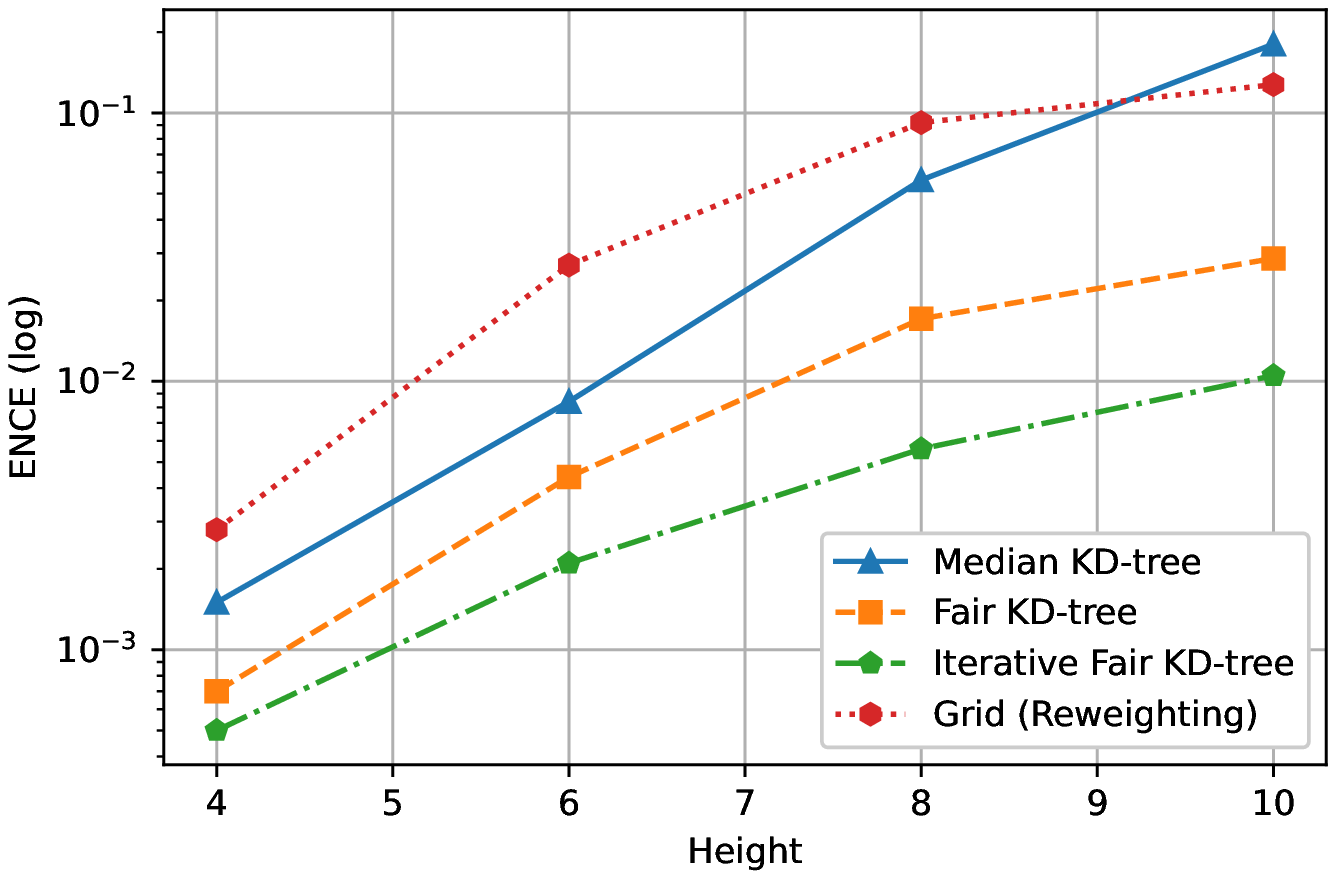}
	}
	\hfill
	\subfloat[Houston (Decision Tree)\label{figure: disparity proof s8}]{%
	\includegraphics[scale = 0.35]{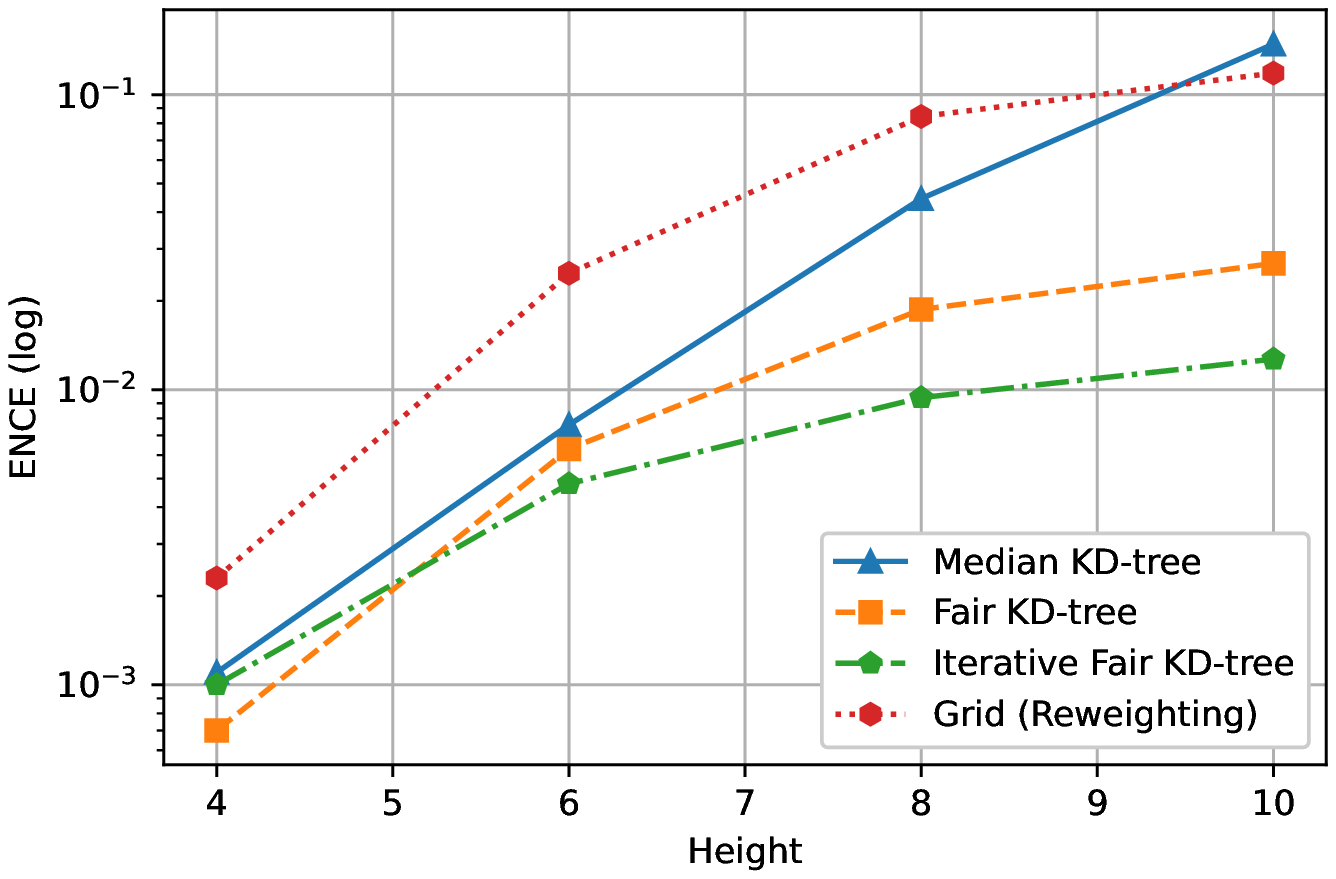}
	}
	\hfill
	\subfloat[Houston (Naive Bayes)\label{figure: disparity proof s9}]{%
	\includegraphics[scale = 0.35]{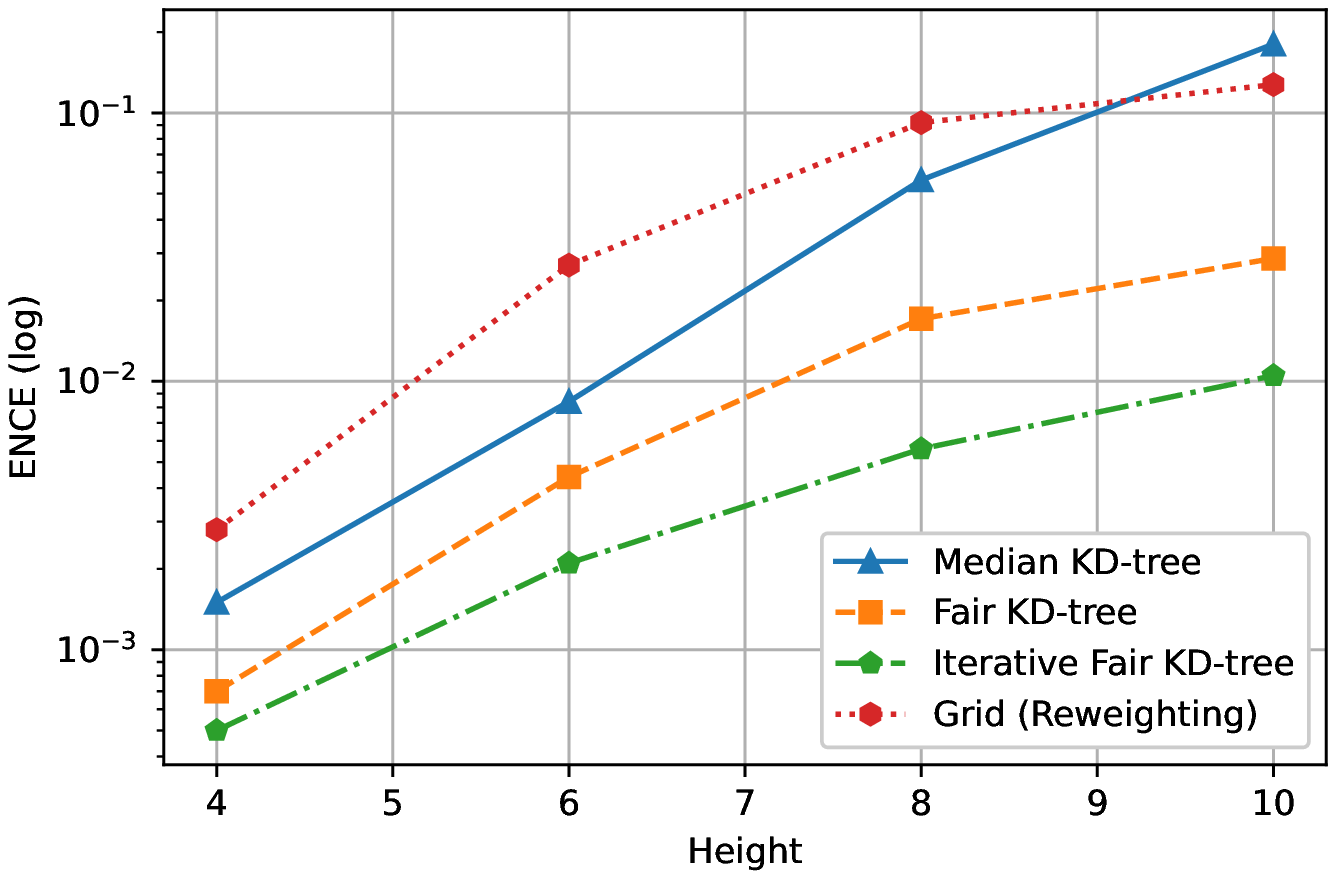}
	}
	\vspace{-10pt}
	\caption{Performance Evaluation with respect to ENCE.}
    \label{figure: ENCE}
	\vspace{-10pt}
\end{figure*}

\subsection{Experimental Setup}\label{section: experimental setup}

We use two real-world datasets provided by EdGap~\cite{edgap}
with $1153$ and $966$ data records respectively, containing socio-economic features (e.g., household income and family structure) of US high school students in Los Angeles, CA and Houston, Texas.
Consistent with~\cite{fischer_2021}, we use two features of average American College Testing (ACT) and the percentage of family employment as indicators to generate classification labels. The geospatial coordinates of schools are derived by linking their identification number to data provided by the National Center for Education Statistics~\cite{education}.

We evaluate the performance of our proposed approaches (Fair KD-tree, Iterative Fair KD-tree, and multi-objective Fair KD-tree) in comparison with three benchmarks: (i) Median KD-tree, the standard method for KD-tree partitioning; (ii) Reweighting over grid -- an adaptation of the re-weighting approach used in in~\cite{kamiran2012data} and deployed in geospatial tools such as {\em IBM AI Fairness 360};
and (iii) we use zip code partitioning as a baseline partitioning. All experiments are implemented in Python and executed on a $3.40$GHz core-i7 Intel processor with 16GB RAM.


\begin{figure*}[tbh]
	\subfloat[Model Accuracy (Los Angeles)\label{figure: wac single task s2}]{%
	\includegraphics[scale=.35]{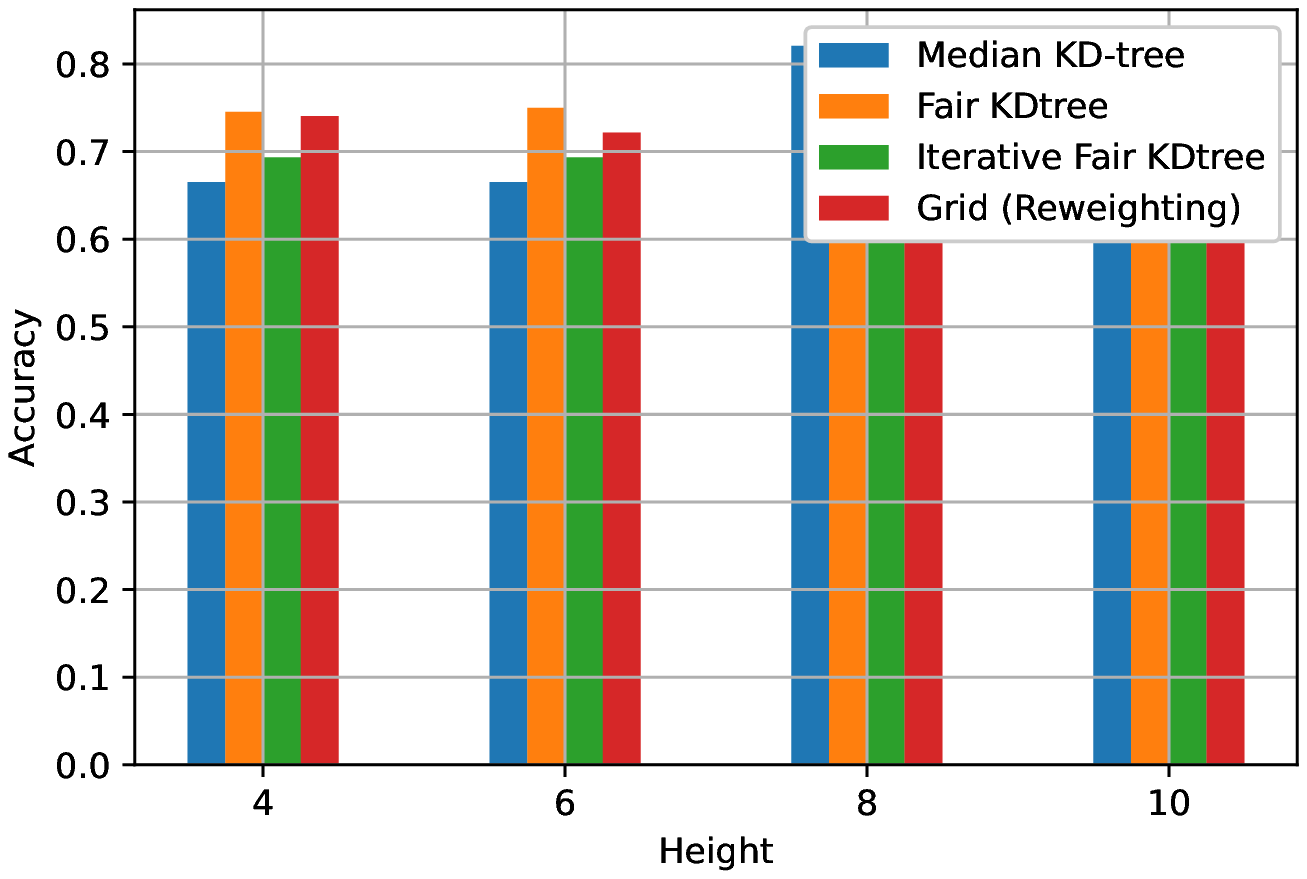}
	}
	\hfill
	\subfloat[Training Miscalibration (Los Angeles)\label{figure: wac single task s3}]{%
	\includegraphics[scale=.35]{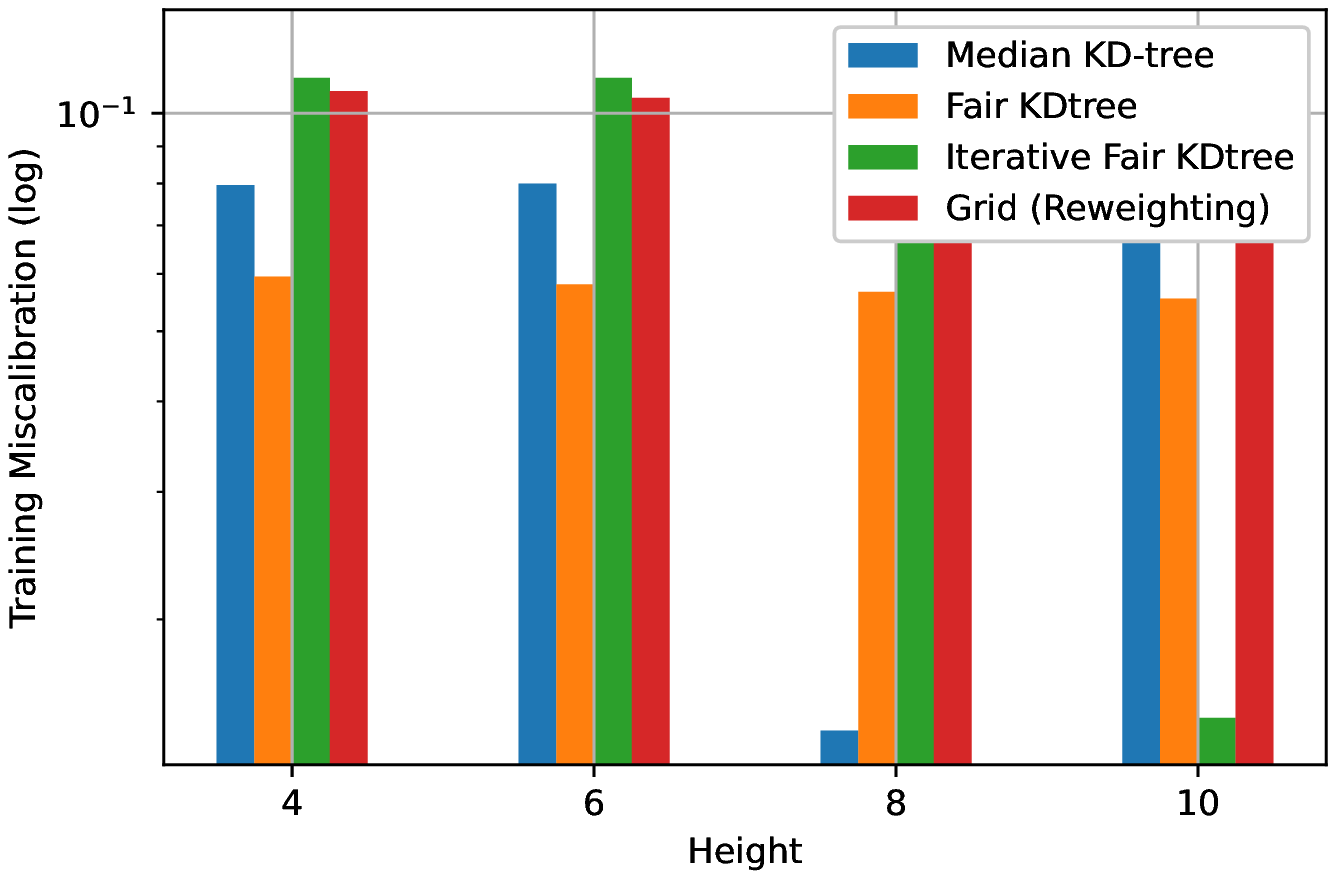}
	}
	\hfill
	\subfloat[Test Miscalibration (Los Angeles)\label{figure: wac single task s4}]{%
	\includegraphics[scale=.35]{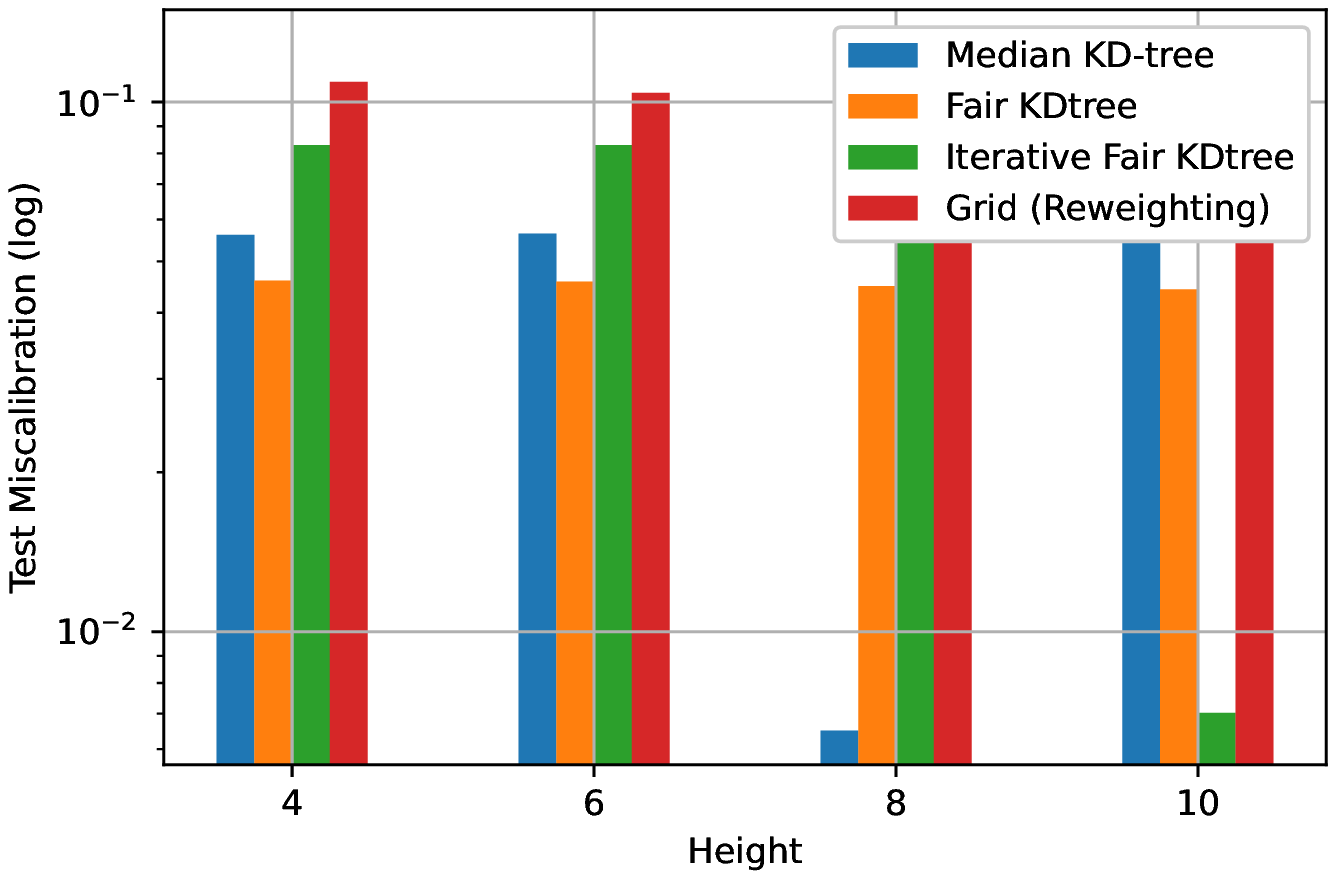}
	}
	\hfill
	\subfloat[Model Accuracy (Houston)\label{figure: wac single task s6}]{%
	\includegraphics[scale=.35]{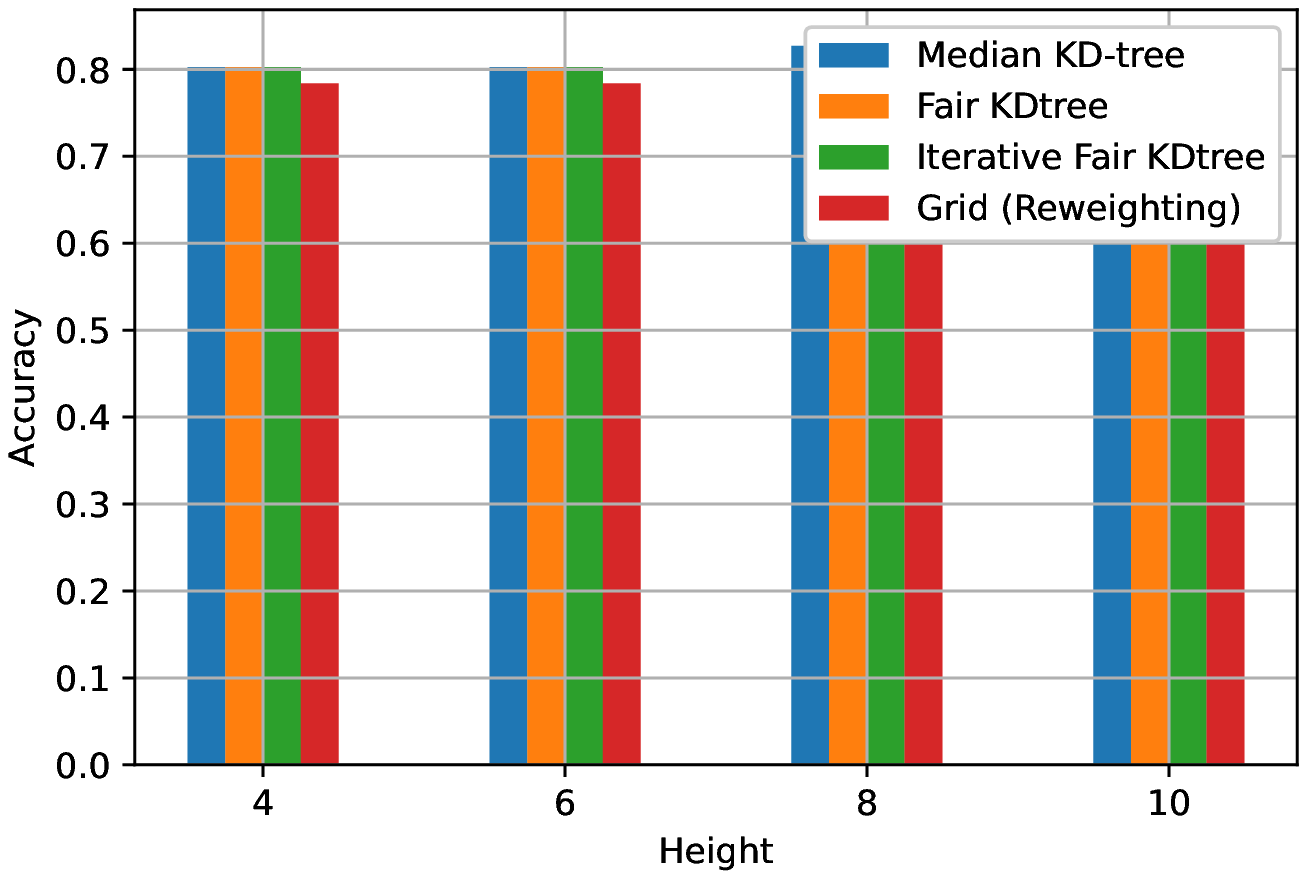}
	}
	\hfill
	\subfloat[Training Miscalibration (Houston)\label{figure: wac single task s7}]{%
	\includegraphics[scale=.35]{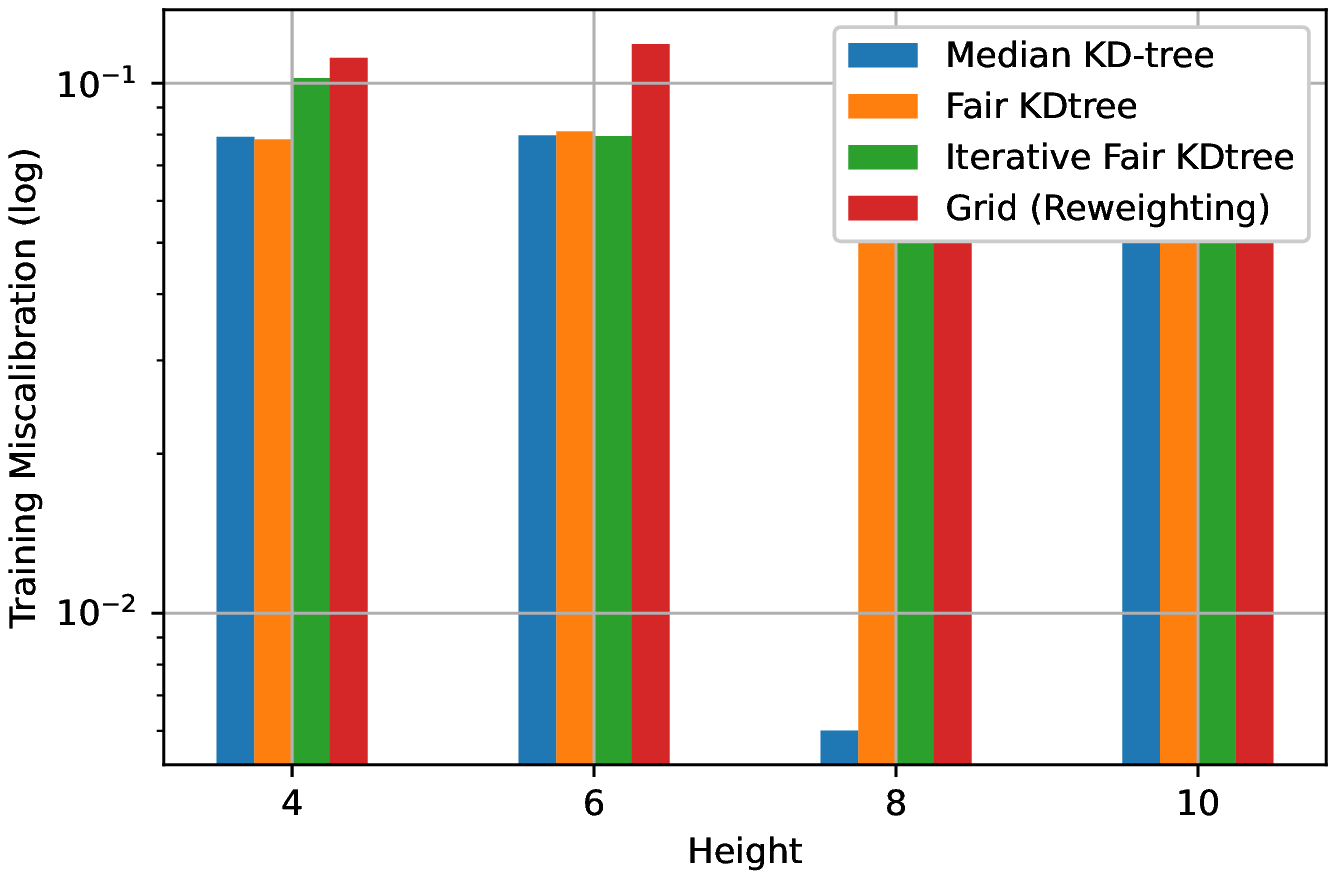}
	}
	\hfill
	\subfloat[Test Miscalibration (Houston)\label{figure: wac single task s8}]{%
	\includegraphics[scale=.35]{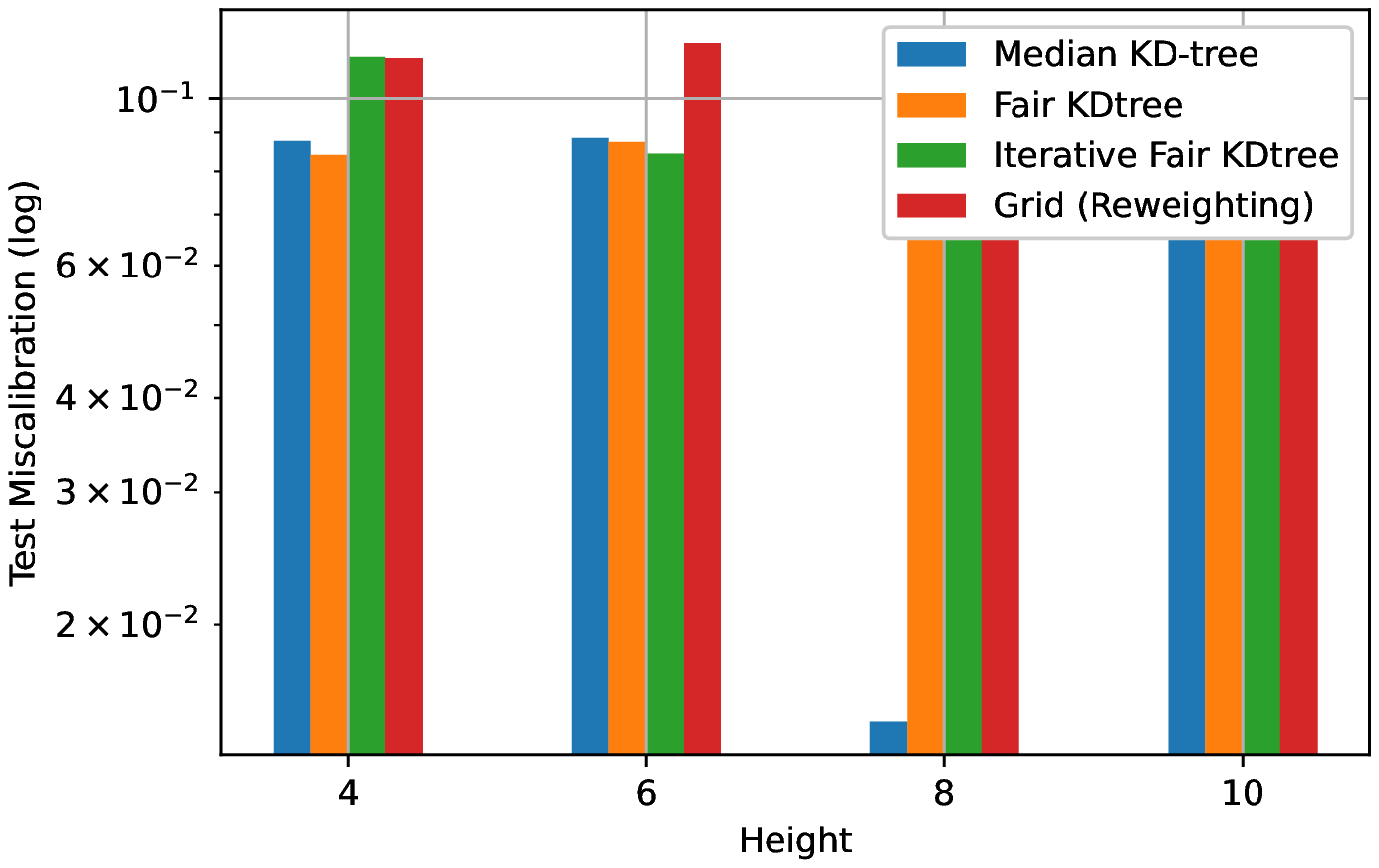}
	}
	\vspace{-10pt}
	\caption{Performance Evaluation with respect to other indicators.}
    \label{figure: other indicators}
	\vspace{-10pt}
\end{figure*}

\subsection{Evidence for Disparity in Geospatial ML}\label{section: diparity}

First, we perform a set of experiments to measure the amount of bias that occurs when performing ML on geospatial datasets without any mitigating factors. Figure~\ref{figure: disparity proof} captures the existing disparity with respect to widely accepted metrics of  calibration error and ECE with $15$ bins. We use the ratio representation of calibration in which a closer value to $1$ represents higher calibration levels. Two logistic regression models are trained over neighborhoods in Los Angeles and Houston areas. The labels are generated by setting a threshold of $22$ on the average ACT performance of students in schools. The overall performance of models in terms of 
training and test calibration in Los Angeles and Texas are $(1.005,\, 1.033)$ and $(0.999,\, 0.958)$, respectively.
Both training and test calibration are close to $1$ overall, which in a naive interpretation would indicate all schools are treated fairly. However, this is not the case when computing the same metrics on a per-neighborhood basis. Figure~\ref{figure: disparity proof} shows miscaibration error for the top $10$ most populated zip codes. Despite the model's acceptable outcomes {\em overall}, many {\em individual} neighborhoods suffer from severe calibration errors, leading to unfair outcomes in the most populated regions, which are often home to the under-privileged communities.




\begin{figure*}[t]
	\subfloat[Median KD-tree (Los Angeles)\label{figure: heatmap s1}]{%
	\includegraphics[scale=.35]{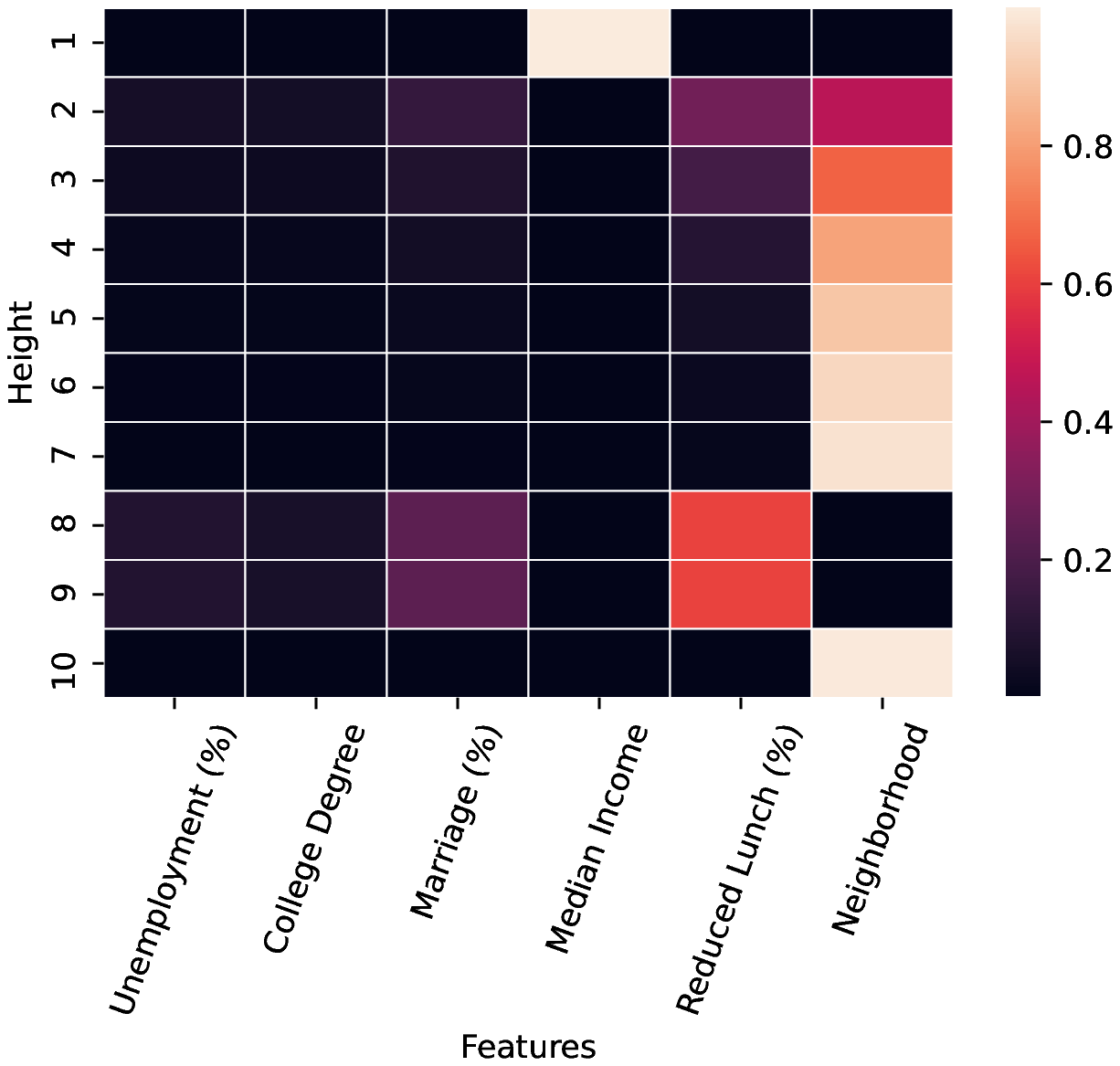}
	}
	\hfill
	\subfloat[Fair KD-tree  (Los Angeles)\label{figure: heatmap s2}]{%
	\includegraphics[scale=.35]{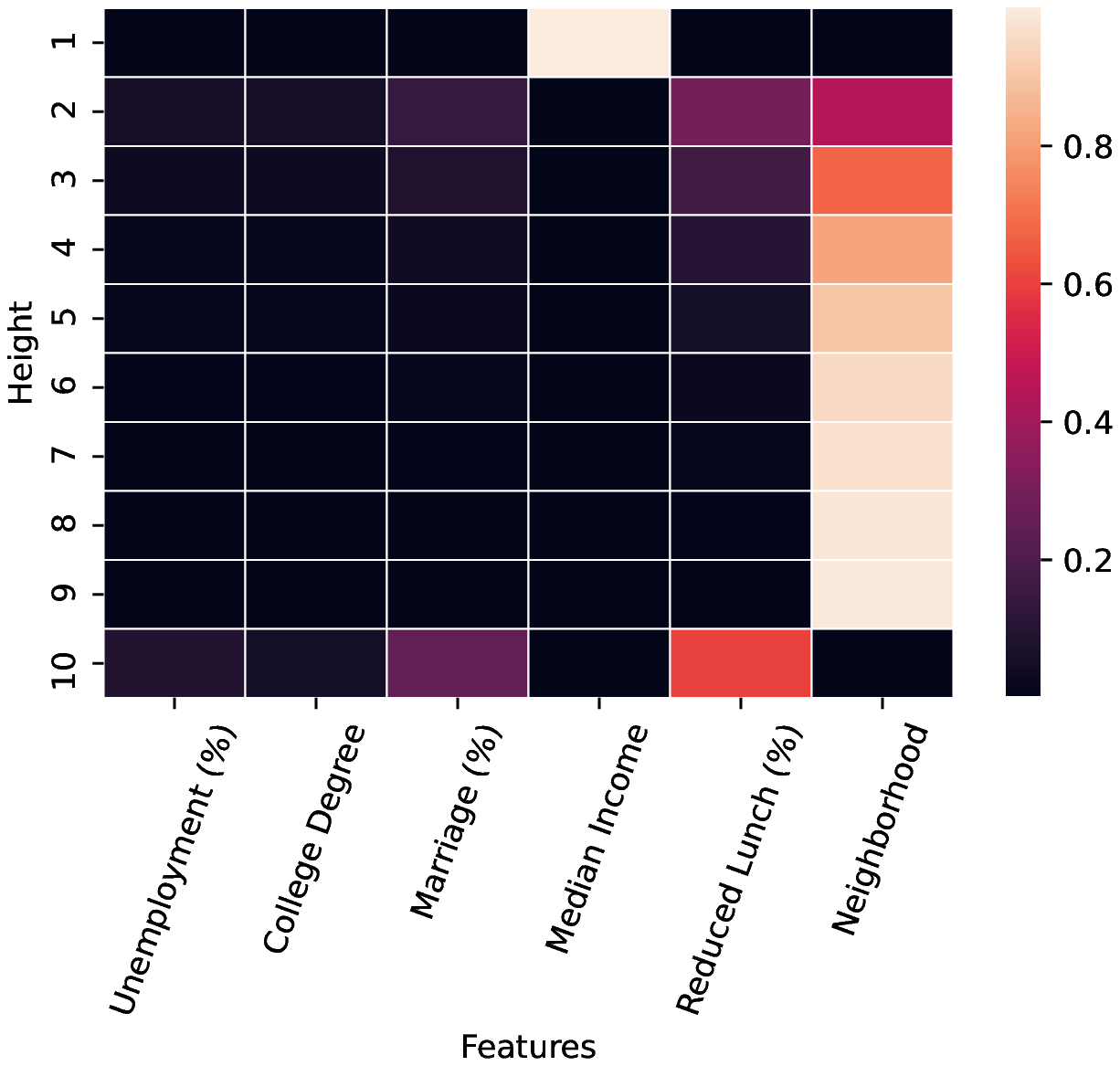}
	}
	\hfill
	\subfloat[Iterative Fair KD-tree (Los Angeles)\label{figure: heatmap s3}]{%
	\includegraphics[scale=.35]{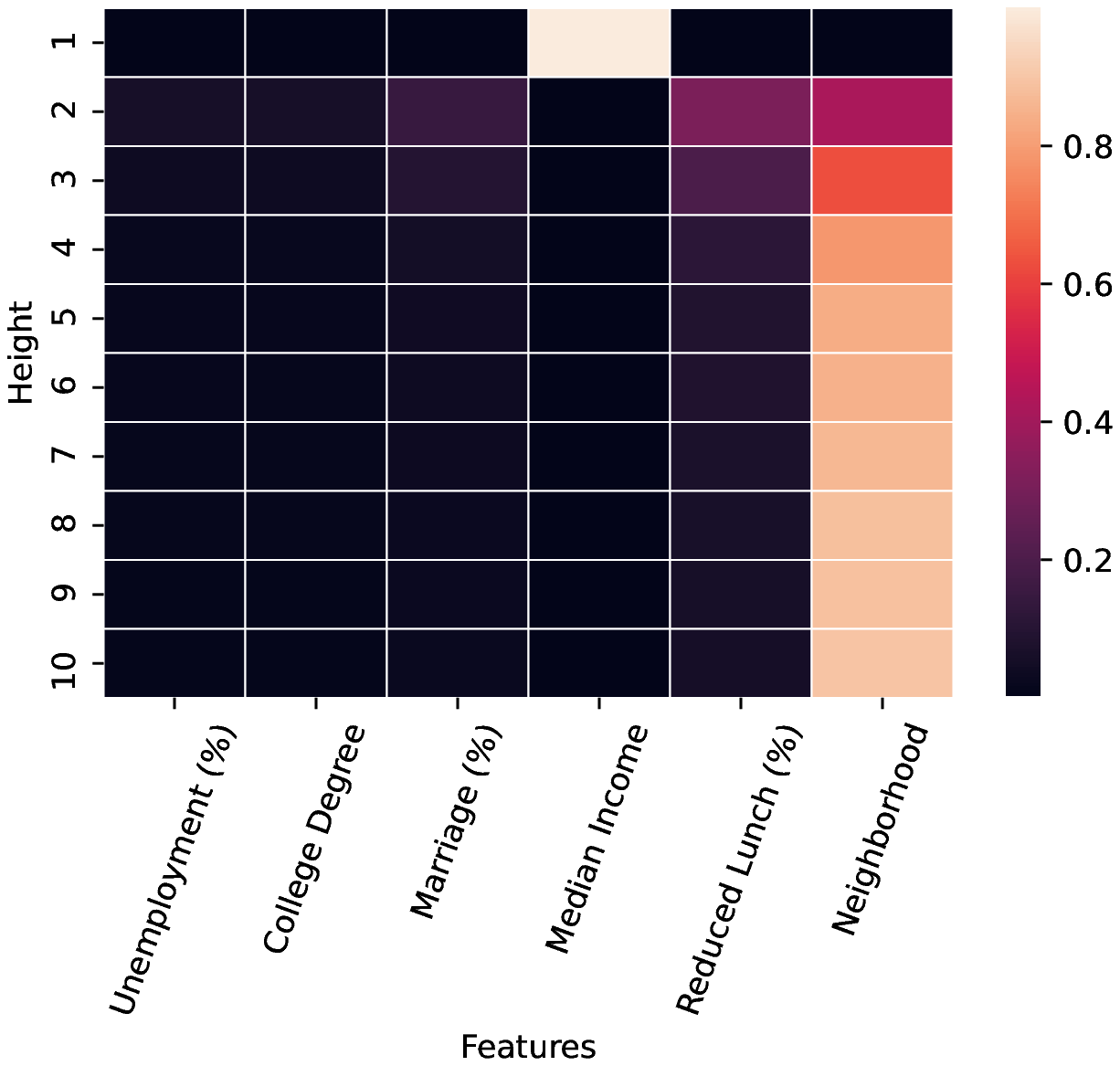}
	}
    \hfill
	\subfloat[Median KD-tree (Houston)\label{figure: heatmap s4}]{%
	\includegraphics[scale=.35]{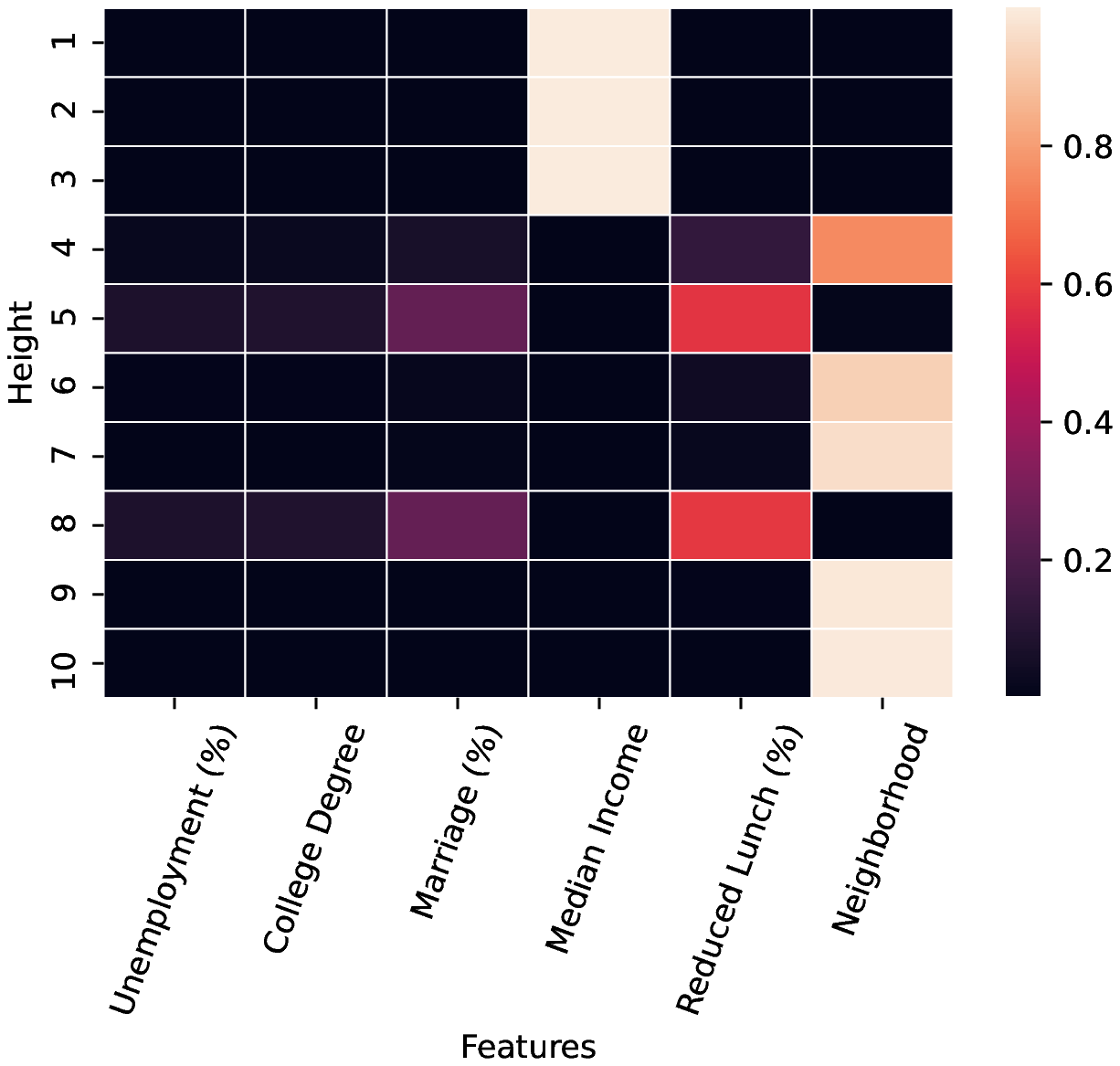}
	}
	\hfill
	\subfloat[Fair KD-tree  (Houston)\label{figure: heatmap s5}]{%
	\includegraphics[scale=.35]{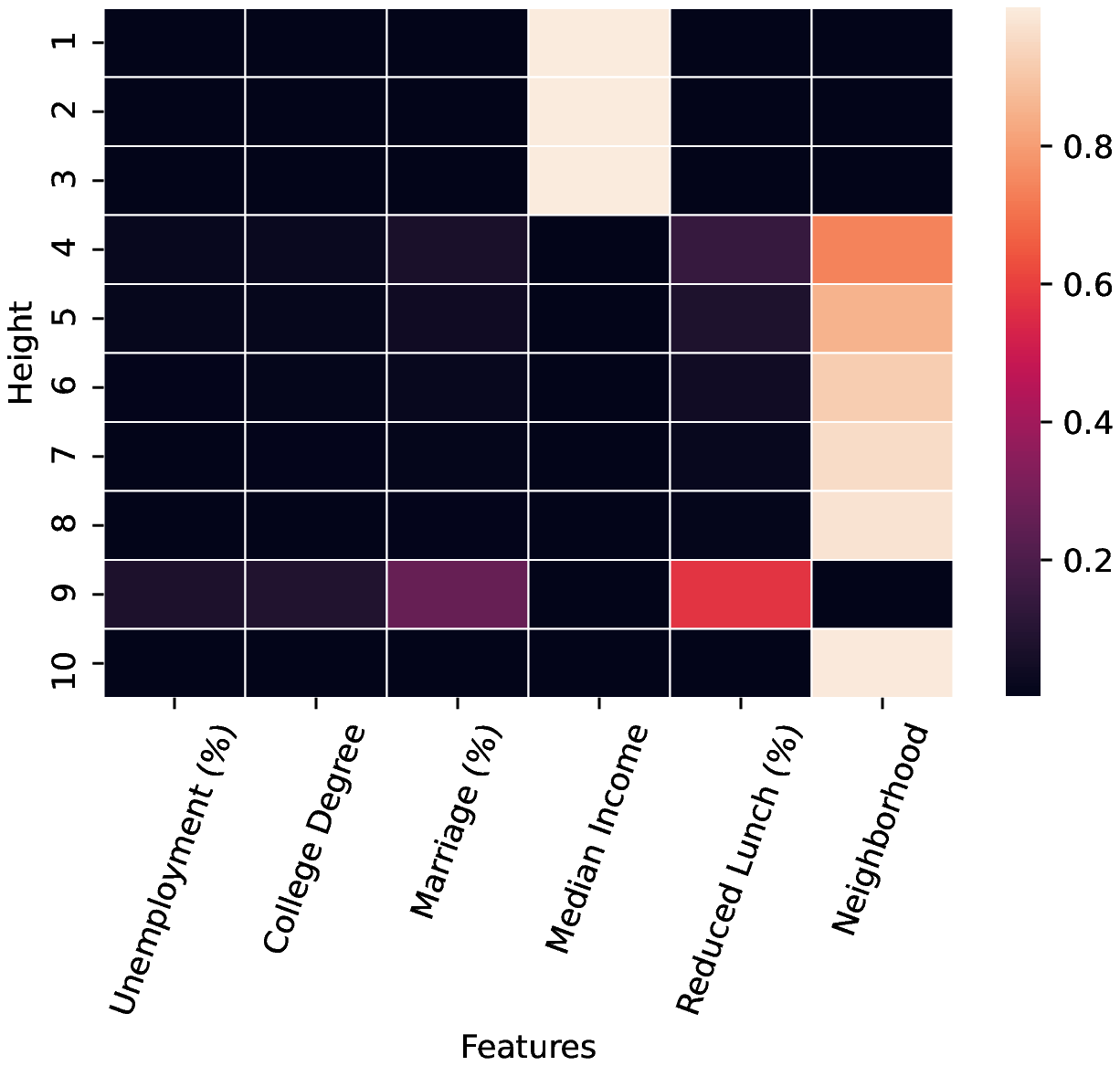}
	}
	\hfill
	\subfloat[Iterative Fair KD-tree  (Houston)\label{figure: heatmap s6}]{%
	\includegraphics[scale=.35]{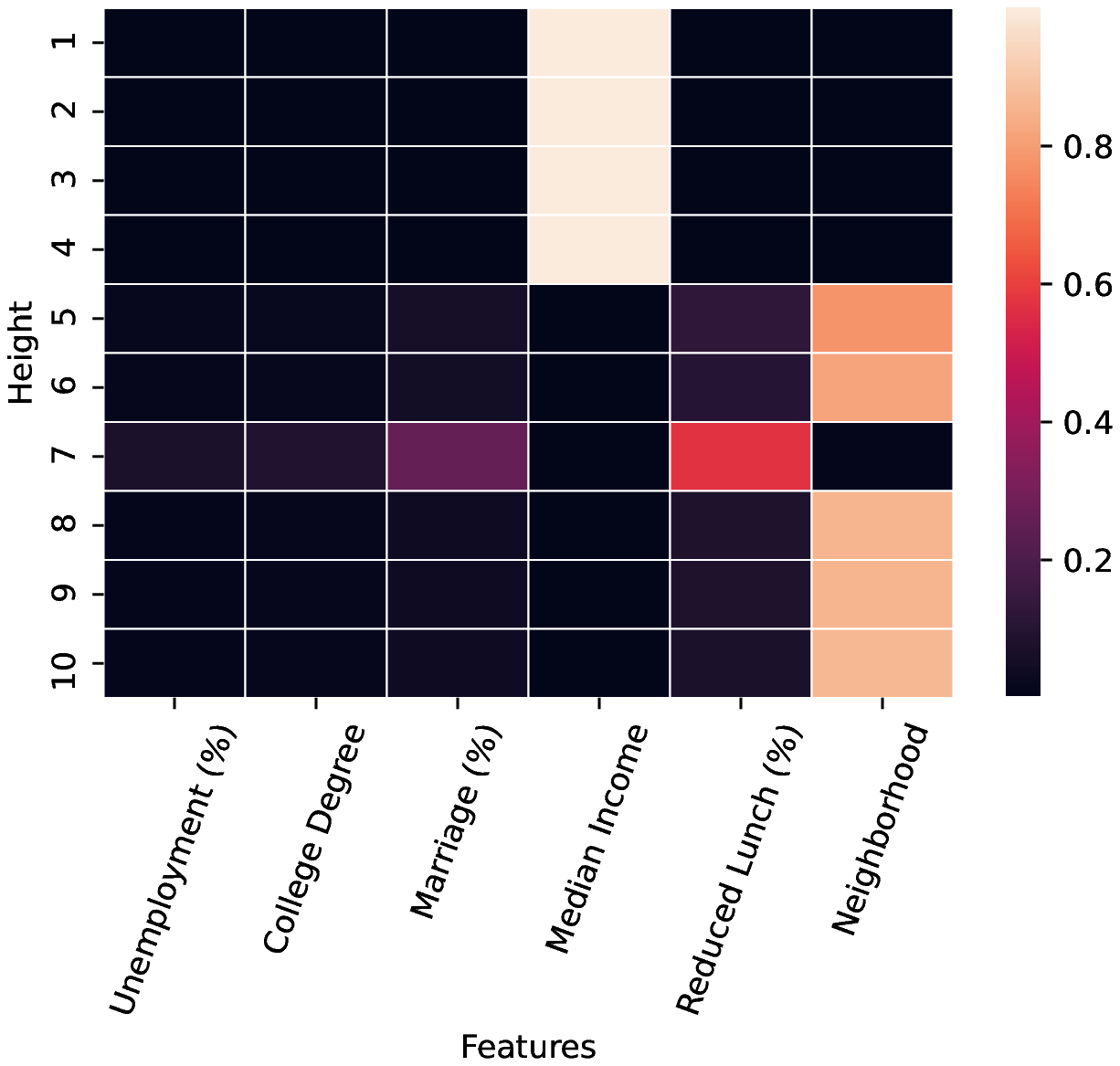}
	}
	\vspace{-10pt}
	\caption{Impact of features on decision-making.}
    \label{figure: heatmap}
	\vspace{-10pt}
\end{figure*}

\vspace{-10pt}

\subsection{Mitigation Algorithms}\label{section: performance analysis}


\subsubsection{Evaluation w.r.t. ENCE Metric.}
ENCE is our primary evaluation metric that captures the amount of calibration error over neighborhoods. Recall that Fair KD-tree and its extension Iterative Fair KD-tree can work for any given classification ML model. We apply algorithms for Logistic Regression, Decision Tree, and Naive Bayes classifiers to ensure diversity in models. We focus on student SAT performance following the prior work in~\cite{fischer_2021} by setting the threshold to $22$ for label generation. Figure~\ref{figure: ENCE} provides the results in Los Angeles and Houston on the EdGap dataset. The $x$-axis denotes the tree's height used in the algorithm. Having a higher height  indicates a finer-grained partitioning. The $y$-axis is log-scale.



Figure~\ref{figure: ENCE} demonstrates that both Fair KD-tree and Iterative Fair KD-tree outperform benchmarks by a significant margin. The improvement percentage increases as the number of neighborhoods increase, which is an advantage of our techniques, since finer spatial granularity is beneficial for most analysis tasks. The intuition behind this trend lies in the overall calibration of the model: given that the trained model is well-calibrated overall, dividing the space into a smaller number of neighborhoods is expected to achieve a calibration error closer to the overall model. 
This result supports Theorem~\ref{Theorem: calibration}, stating that ENCE is lower-bounded by the number of neighborhoods. Iterative Fair KD-tree behaves better, as confidence scores are updated on every tree level. The improvement achieved compared to Fair KD-trees comes at the expense of higher computational complexity. On average Fair KD-tree achieves $45 \%$ better performance in terms of computational complexity. The time taken for Fair KD-tree with 10 levels is $102$ seconds, versus $189$ seconds for the iterative version.


\begin{figure*}[tbh]
	\subfloat[Height$=4$, Los Angeles \label{figure: multi task s1} ]{%
	\includegraphics[scale=.3]{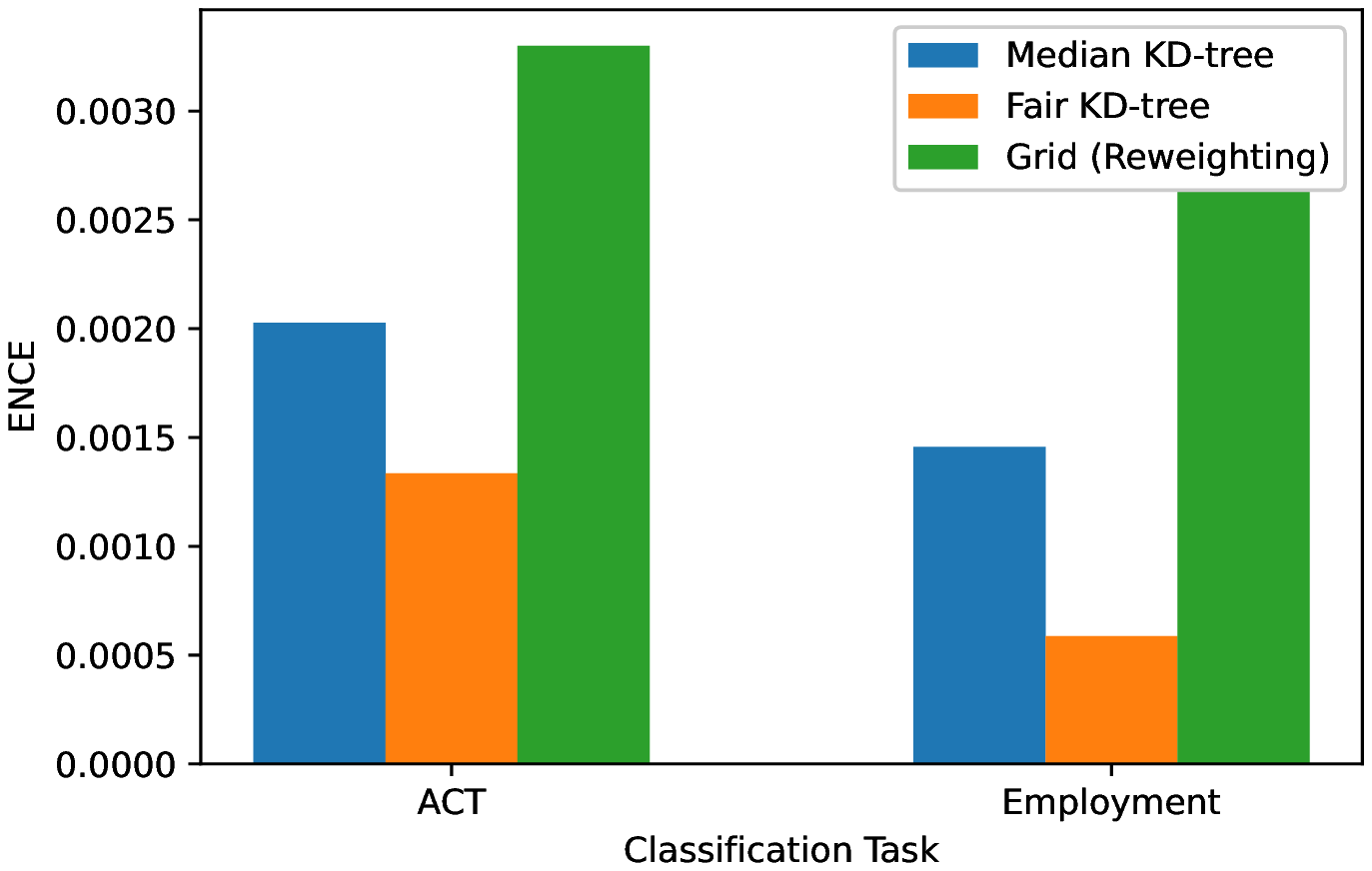}
	}
	\hfill
	\subfloat[Height$=6$, Los Angeles \label{figure: multi task s2} ]{%
	\includegraphics[scale=.3]{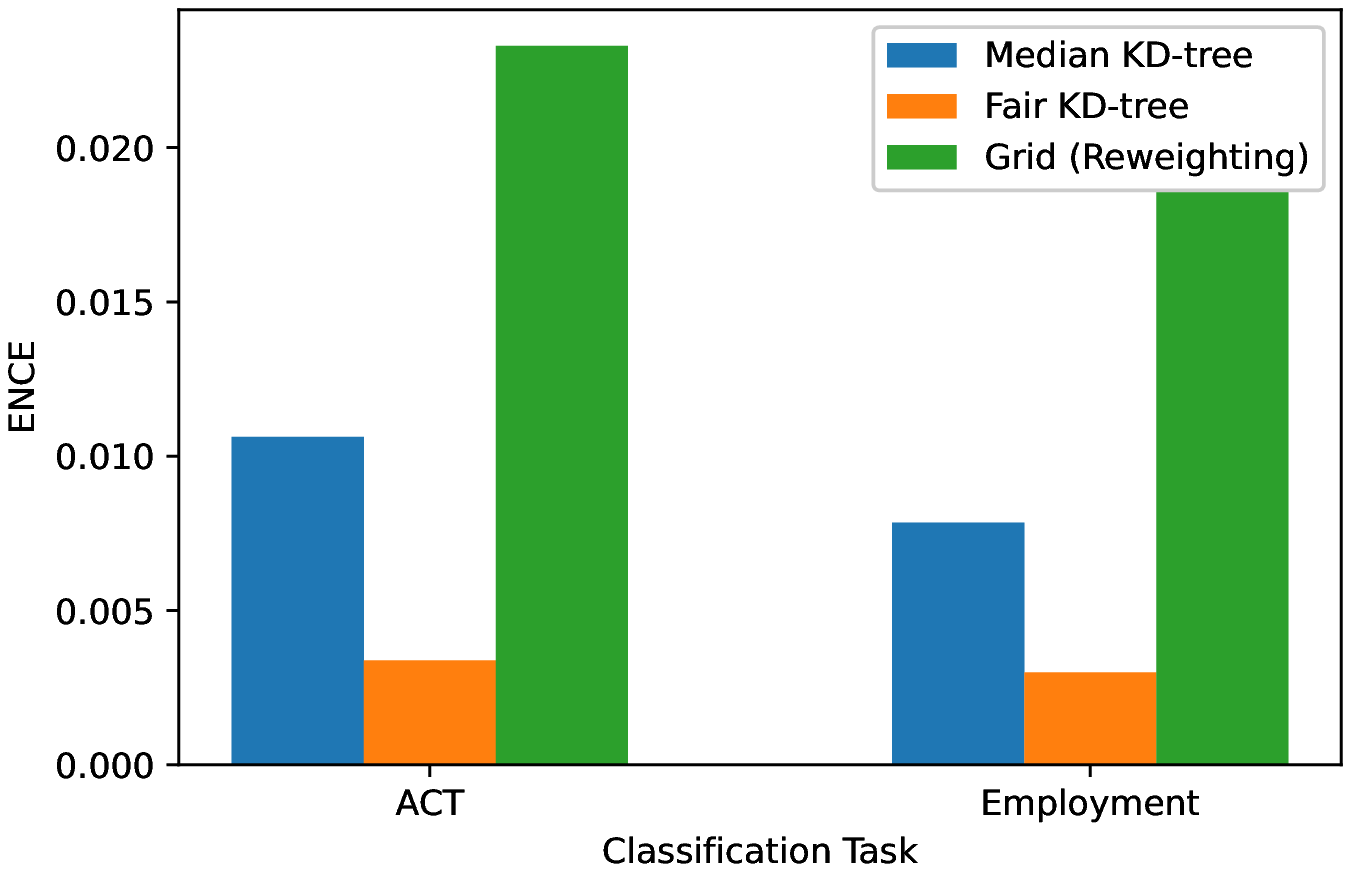}
	}
	\hfill
	\subfloat[Height$=8$, Los Angeles \label{figure: multi task s3} ]{%
	\includegraphics[scale=.3]{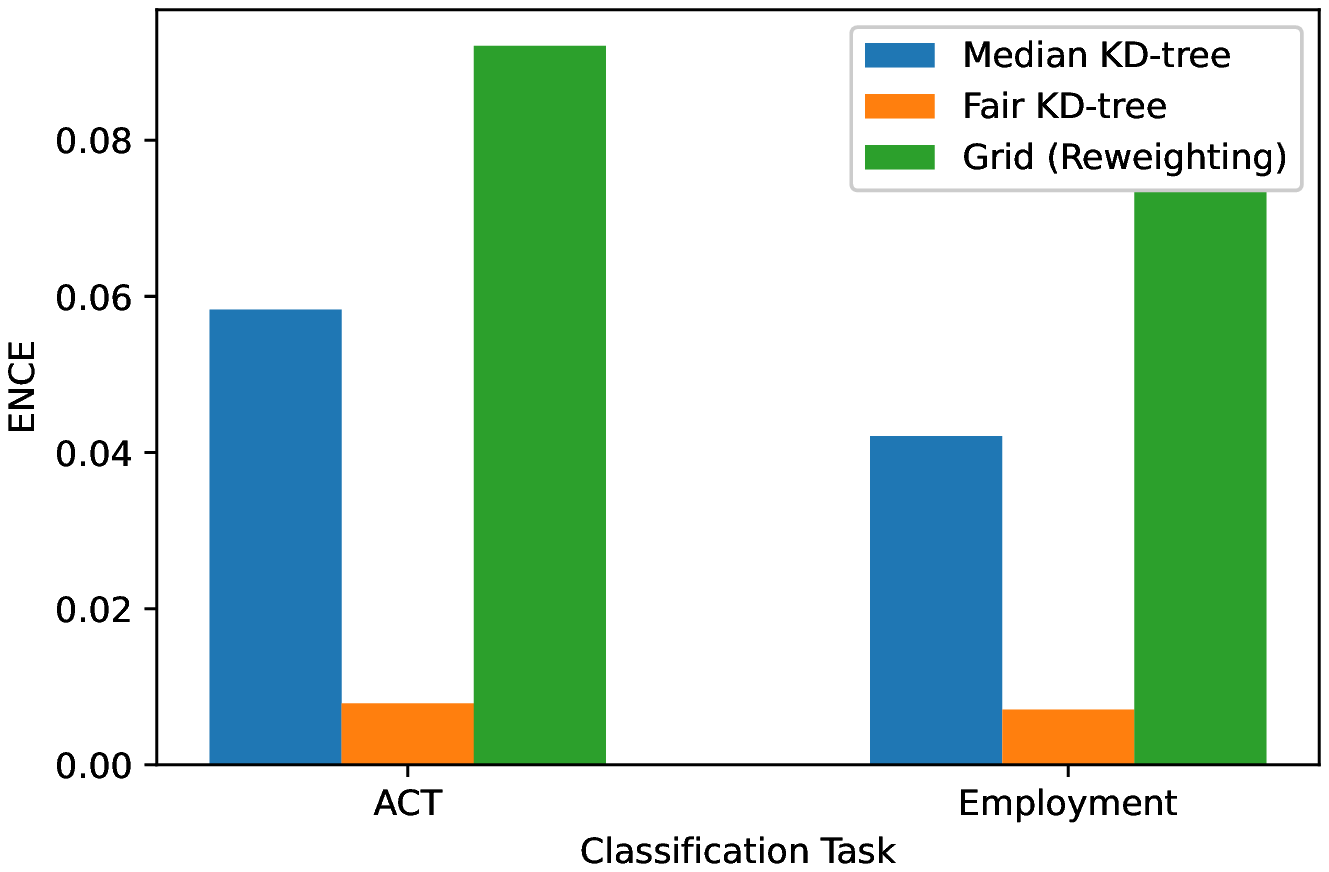}
	}
	\hfill
	\subfloat[Height$=10$, Los Angeles \label{figure: multi task s4} ]{%
	\includegraphics[scale=.3]{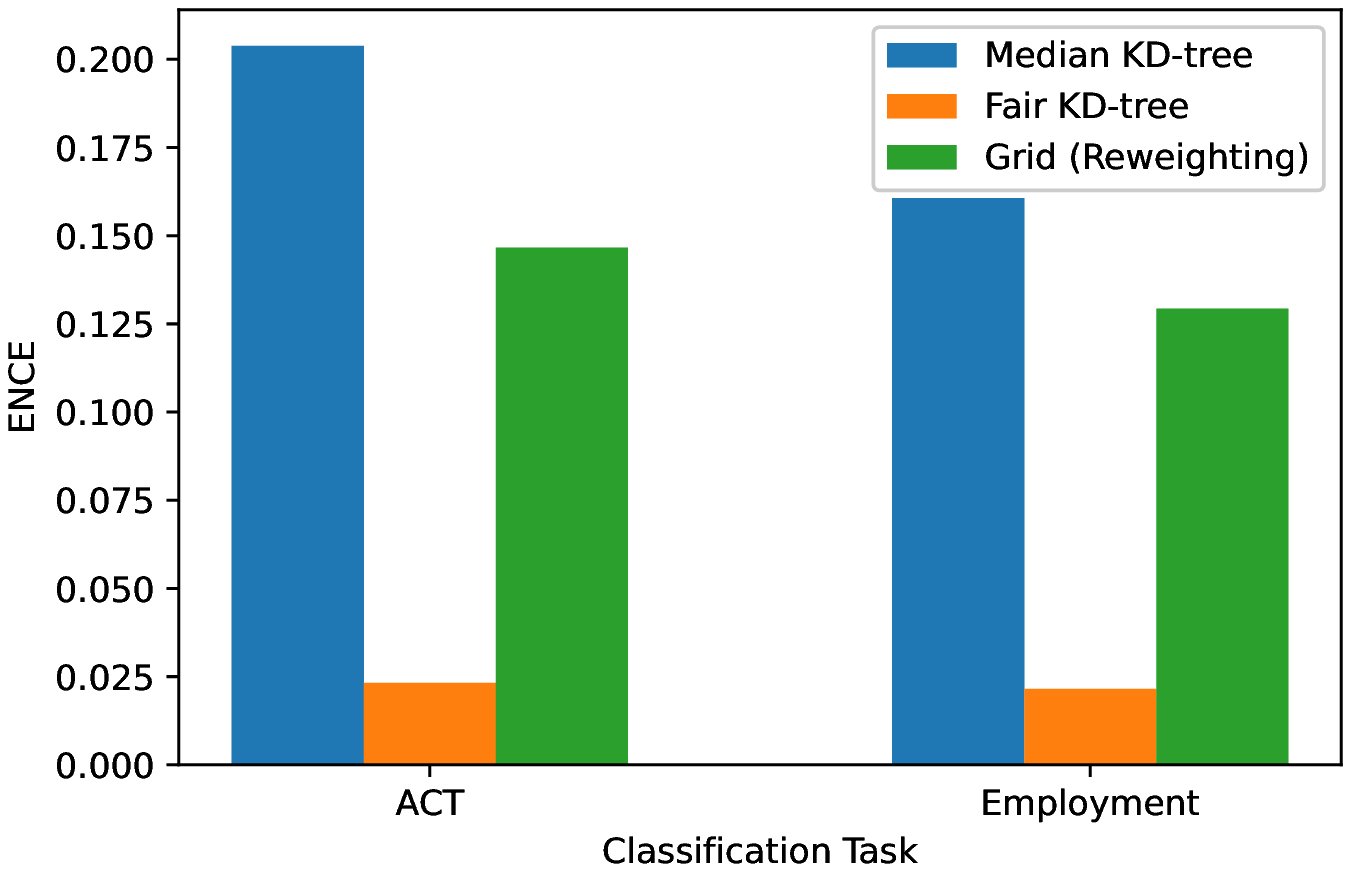}
	}
	\hfill
	\subfloat[Height$=4$, Houston\label{figure: multi task s5} ]{%
	\includegraphics[scale=.3]{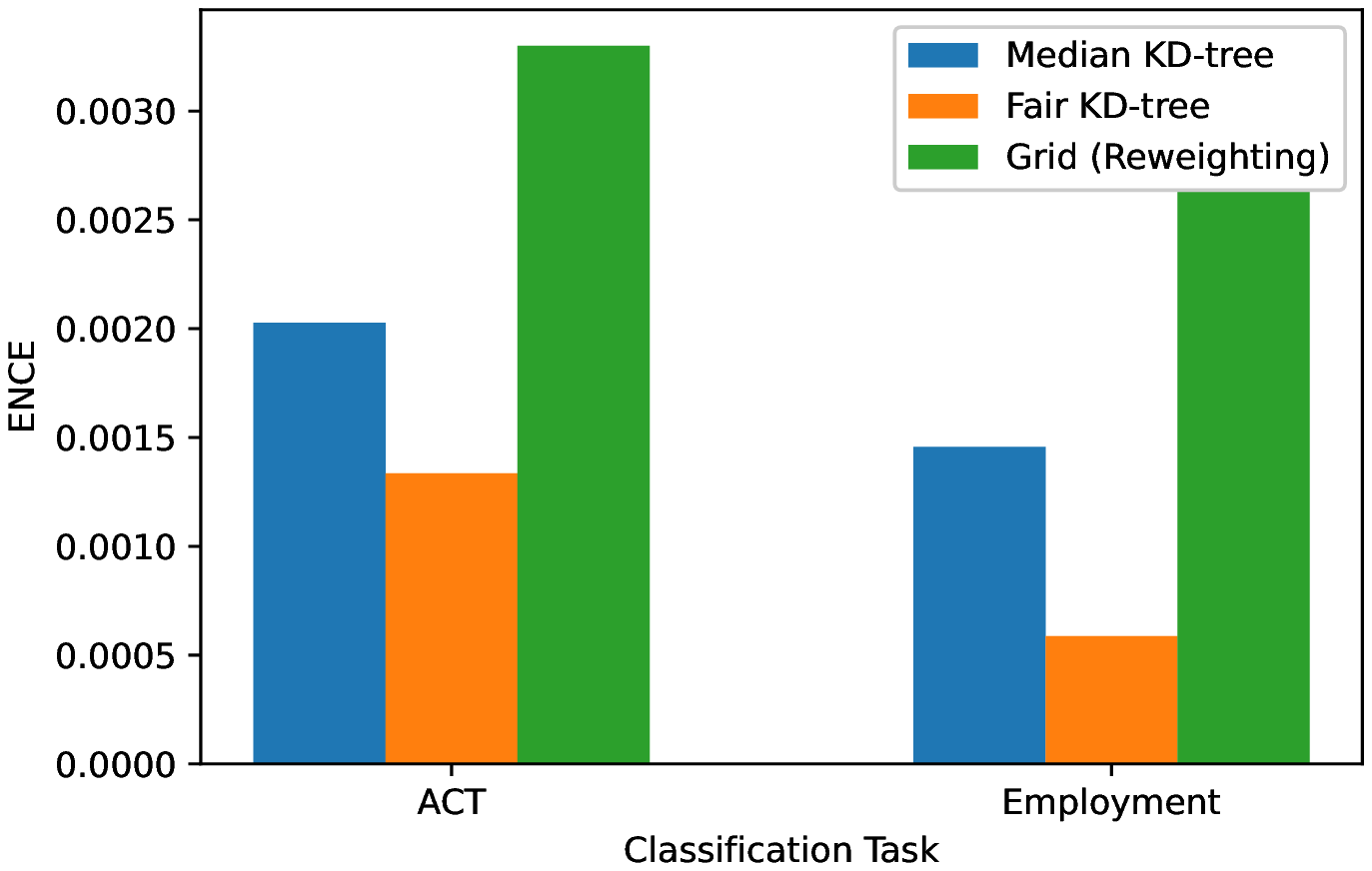}
	}
	\hfill
	\subfloat[Height$=6$, Houston\label{figure: multi task s6} ]{%
	\includegraphics[scale=.3]{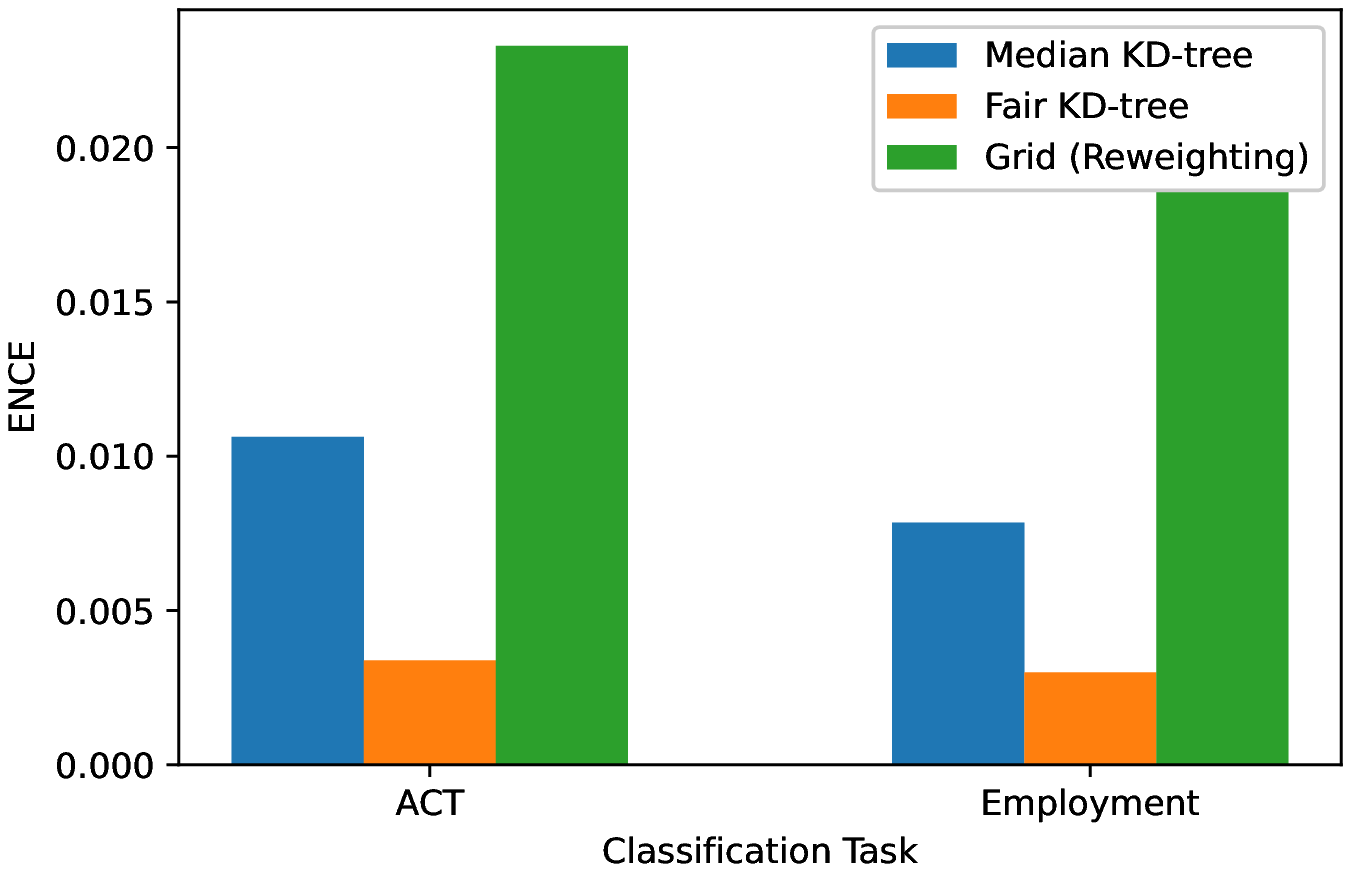}
	}
	\hfill
	\subfloat[Height$=8$, Houston\label{figure: multi task s7} ]{%
	\includegraphics[scale=.3]{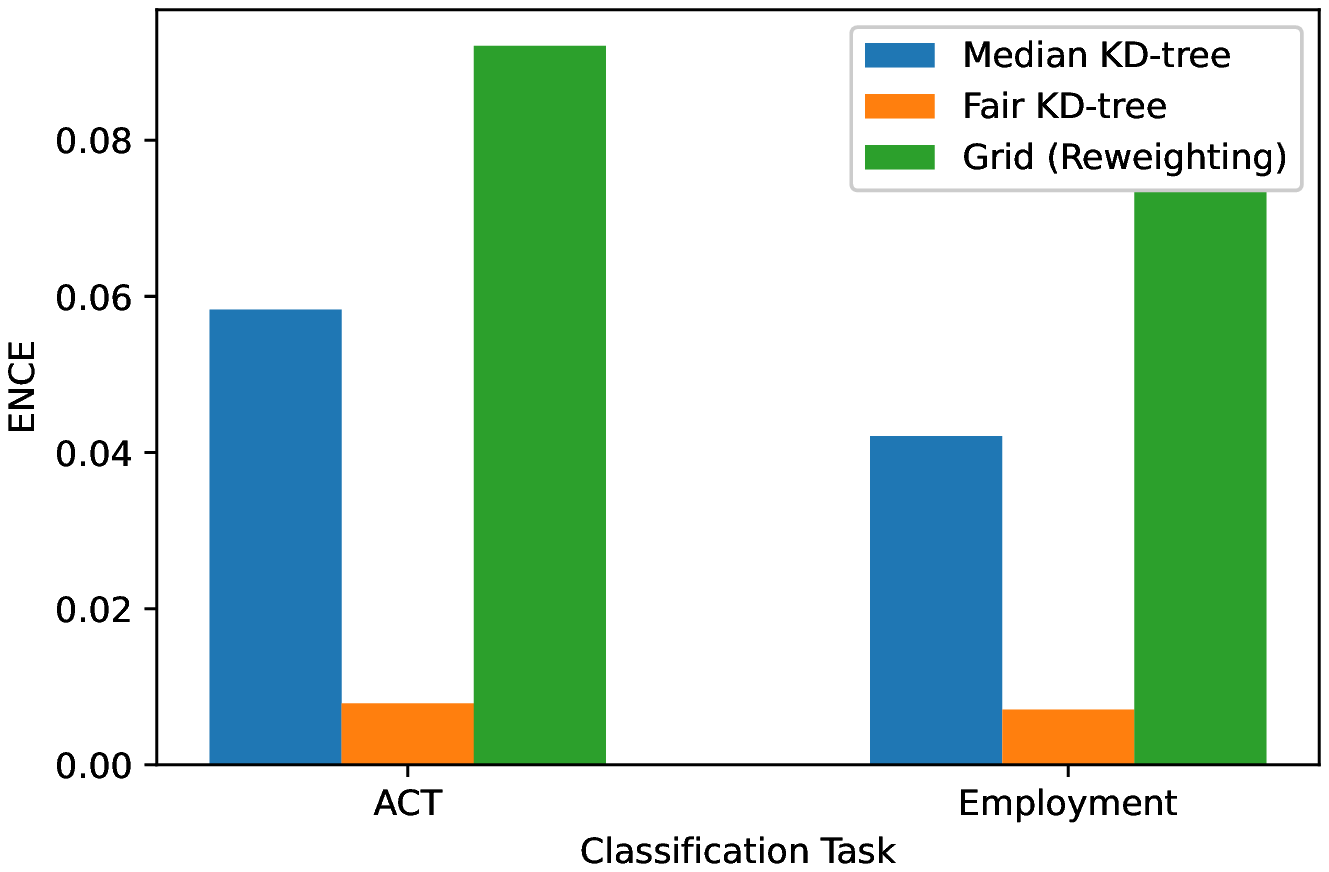}
	}
	\hfill
	\subfloat[Height$=10$, Houston\label{figure: multi task s8} ]{%
	\includegraphics[scale=.3]{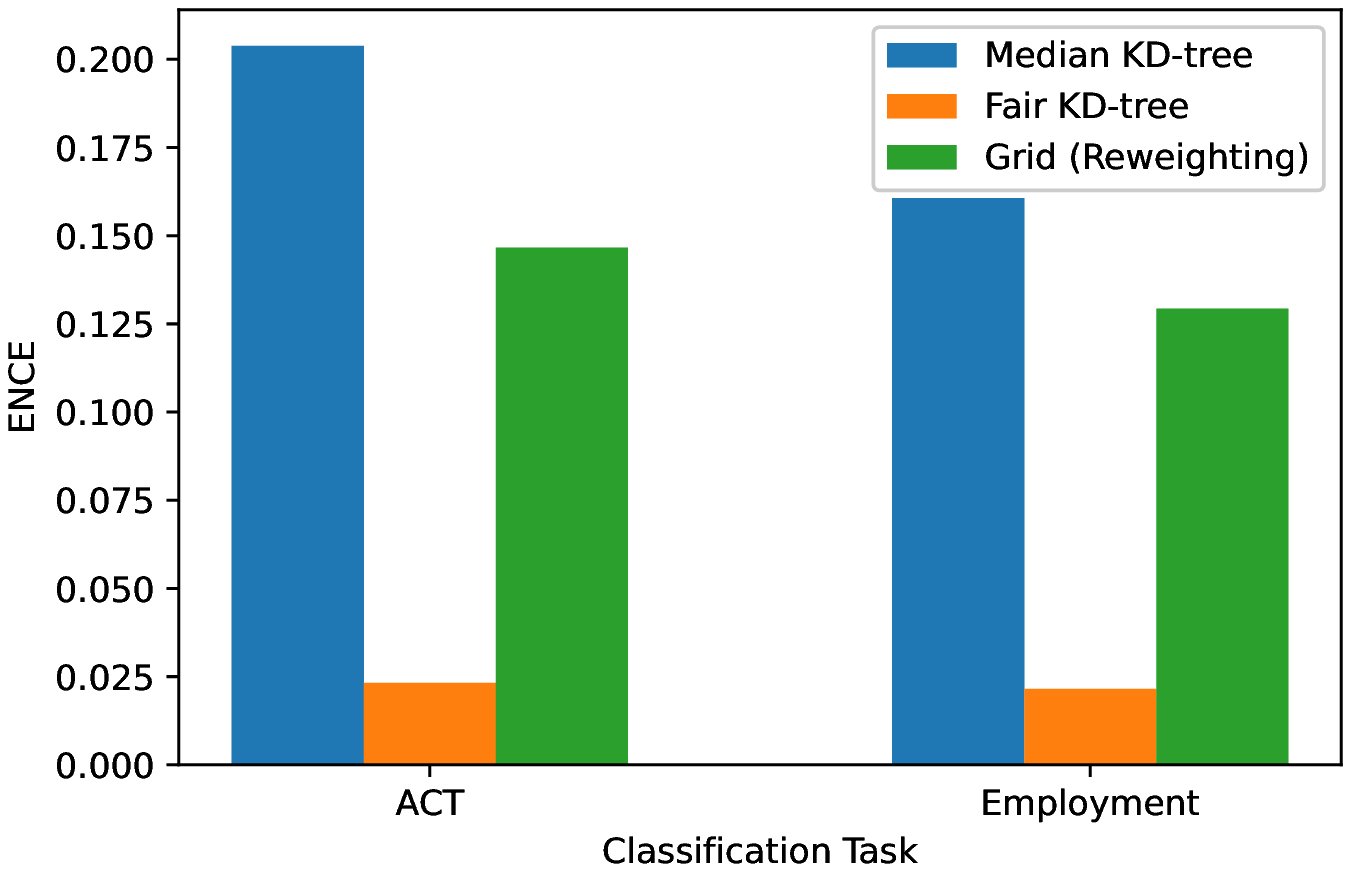}
	}
	\vspace{-10pt}
	\caption{Performance evaluation of multi-objective algorithm.}
    \label{figure: multi task} 
	\vspace{-10pt}
\end{figure*}

\subsubsection{Evaluation w.r.t. other Indicators.}
In Figure~\ref{figure: other indicators} we evaluate fairness with respect to three other key indicators:  model accuracy, training miscalibration, and test miscalibration. We focus on logistic regression, one of the most widely adopted classification units. The accuracy of all algorithms follows a similar pattern and increases at higher tree heights. This is expected, as more geospatial information can be extracted at finer granularities. 

Figure~\ref{figure: wac single task s3} shows training miscalibration calculated for the overall model (a lower value of calibration error indicates better performance). Our proposed algorithms have comparable calibration errors to benchmarks, even  though their fairness is far superior. To understand better underlying trends, Figure~\ref{figure: heatmap} provides the heatmap for the tree-based algorithms over $10$ different tree heights. The amount of contribution each feature has on decision-masking is captured using a different color code. One observation is that the model shifts focus to different features based on the height. Such sudden changes can impact the generated confidence scores and, subsequently, the overall calibration of the model. As an example, consider the median KD-tree algorithm at the height of $8$ in Los Angeles (Figure~\ref{figure: wac single task s3}): there is a sudden drop in training calibration, which can be explained by looking at the corresponding heat map in Figure~\ref{figure: heatmap s1}. At the height of $8$, the influential features on decision-making consist of different elements than the heights $4$, $6$, and $10$, leading to the fluctuation in the model calibration.

\subsection{Performance of multi-objective approach.}\label{section: multi-objective}

When multi-objective criteria are used, we need a methodology to unify the geospatial boundaries generated by each task. Our proposed multi-objective fair partitioning predicated on Fair KD-trees addresses exactly this problem. In our experiments, we use the two criteria of ACT scores and employment percentage of families as the two objectives used for partitioning. These features are separated from the training dataset in the pre-processing phase and are used to generate labels. The threshold for ACT is selected as before ($22$), and the threshold for label generation based on family employment is set to $10$ percent.

Figure~\ref{figure: multi task} presents the results of the Multi-Objective Fair KD-tree (to simplify chart notation, we use the `Fair KD-tree' label). We choose a $\alpha$ value of $0.5$ to give equal weight to both objectives. We emphasize that, the output of the Multi-Objective Fair KD-tree is a single non-overlapping partitioning of the space representing neighborhoods. Once the neighborhoods are generated, we show the performance with respect to each objective function, i.e., ACT and employment. The first row of the figure shows the performance for varying tree heights in Los Angeles, and the second row corresponds to Houston. The proposed algorithm improves fairness for both objective functions. The margin of improvement increases as the height of the tree increases.

\vspace{-10pt}
\section{Conclusion}\label{Sec: Conclusion}
We proposed an indexing-based technique that achieves spatial group fairness in machine learning. Our technique performs a partitioning of the data domain in a way takes into account not only geographical features, but also calibration error. Extensive evaluation results on real data show that the proposed technique is effective in reducing unfairness when training on location attributes, and also preserves data utility. In future work, we plan to further investigate custom split metrics for fairness-aware spatial indexing that take into account data distribution characteristics. We will also investigate alternative indexing structures, such as R$^+$ trees, that completely cover the data domain and provide superior clustering properties.


\bibliographystyle{ACM-Reference-Format}
\bibliography{sample}
\balance


\appendix
\section{Appendix}\label{Sec: Appendix}



\subsection{Expected Calibration Error}\label{appendix: ecec}

Expected Calibration Error (ECE) is one of the primary metrics used to quantify calibration in ML. According to this metric, the output confidence scores are sorted and partitioned into $M$ bins denoted by $B_1,\, ...., \, B_m$. The associated score for each data instance lies within one of the bins. The ECE metric is then calculated over bins as follows:
 
\begin{align}
\text{ECE} = \sum_{m=1}^M \dfrac{B_m}{n}| \text{o}(B_m) - e(B_m) |
\end{align}

\subsection{Theorem Proofs}

\noindent{\bf Proof of Theorem~\ref{Theorem: calibration}}

The proof follows triangle inequality. The weighted calibration of the model can be written as,
\begin{align}
     \sum_{N_i \in \mathcal{N}} &|N_i| \times | e(h|N= N_i)  - o(h|N= N_i)| =           \\
     &\sum_{N_i \in \mathcal{N}} |N_i| \times |  \dfrac{1}{|N_i|} (\sum_{u\in N_i} s_u )  - \dfrac{1}{|N_i|} (\sum_{u\in N_i} y_u )| = \\         
     &\sum_{N_i \in \mathcal{N}}   |   \sum_{u\in N_i} s_u   - \sum_{u\in N_i} y_u | \geq   |   \sum_{u\in D} s_u   - \sum_{u\in D} y_u |  \\   
     &= |D|\times(|e(h)- o(h)|)
\end{align}

\noindent{\bf Proof of Theorem~\ref{Theorem: calibration 2}}

Since $\mathcal{N}_2$ is a subgroup partitioning of $\mathcal{N}_1$ it can be constructed following step-by-step partitioning of neighborhoods in $\mathcal{N}_1$ into finer granularity ones until reaching $\mathcal{N}_2$. Denote $\mathcal{N}_1$ neighborhoods by $\{ N_1, N_2,..., N_t \}$. Without loss of generality, we show that splitting an arbitrary neighborhood $N_j\in \mathcal{N}_1$ to $N_{j1}$ and $N_{j2}$ leads to a worse ENCE metric value:

\begin{align}
     \text{ENCE}(\mathcal{N}_1) = &\sum_{N_i \in \mathcal{N}} |N_i| \times | e(h|N= N_i)  - o(h|N= N_i)| =           \\
     & \sum_{N_i \in \mathcal{N},i\neq j} |N_i| \times | e(h|N= N_i)  - o(h|N= N_i)| + \nonumber\\ &|N_j|\times  | e(h|N= N_j)  - o(h|N= N_j)|
\end{align}

\noindent Note that,

\begin{align}
     |N_j|\times  &| e(h|N= N_j)  - o(h|N= N_j)| = \\
     &|N_j| \times |  \dfrac{1}{|N_j|} (\sum_{u\in N_j} s_u )  - \dfrac{1}{|N_j|} (\sum_{u\in N_j} y_u )| = \\
      &|   (\sum_{u\in N_j} s_u )  -  (\sum_{u\in N_j} y_u )|  = \\
      &|   (\sum_{u\in N_{j1}} s_u )  -  (\sum_{u\in N_{j1}} y_u )+  (\sum_{u\in N_{j2}} s_u )  -  (\sum_{u\in N_{j2}} y_u )|  \leq \\
      & |   (\sum_{u\in N_{j1}} s_u )  -  (\sum_{u\in N_{j1}} y_u )| +| (\sum_{u\in N_{j2}} s_u )  -  (\sum_{u\in N_{j2}} y_u )| = \\
      &|N_{j1}|\times  |  \dfrac{1}{|N_{j1}|} (\sum_{u\in N_{j1}} s_u )  - \dfrac{1}{|N_{j1}|} (\sum_{u\in N_{j1}} y_u )| + \nonumber \\
      & |N_{j2}| \times | \dfrac{1}{|N_{j2}|}(\sum_{u\in N_{j2}} s_u )  -  \dfrac{1}{|N_{j2}|}(\sum_{u\in N_{j2}} y_u )| 
\end{align}

Therefore, since by further splitting of neighborhoods, ENCE gets worse and as  $\mathcal{N}_2$ can be reconstructed one division at a time from  $\mathcal{N}_1$, one can conclude that 
\begin{equation}
\text{ENCE}(N_1) \leq \text{ENCE}(N_2)
\end{equation}

\noindent{\bf Proof of Theorem~\ref{Theorem: cc fair kd-tree}}

As the tree is binary, there is a maximum of $\lceil \log(t)\rceil$ partitioning levels. At every level of the tree, the fairness objective function is calculated $|D|$ times, with each computation taking a constant time. Therefore, the required number of computations is $\mathcal{O}(|D|\times \lceil \log(t)\rceil)$. Moreover, the algorithm requires an initial run of the model $h$, which depends on what ML model is employed, represented by the computation complexity of $O(h)$ in the total complexity equation. \\

\noindent{\bf Proof of Theorem~\ref{Theorem: cc iterative fair kd-tree}}

Similar to Fair KD-tree, the total number of levels in Iterative Fair KD-tree is $\lceil \log(t)\rceil$ requiring computational complexity of $\mathcal{O}(|D|\times \lceil \log(t)\rceil)$ to obtain the values for fair partitioning. However, in contrast to the Fair KD-tree algorithm, the iterative version requires the execution of the ML model at every height of the tree. The total computational complexity adds up to $\mathcal{O}(|D|\times \lceil \log(t)\rceil)+ \lceil \log(t)\rceil \times \mathcal{O}(h)$.\\

\noindent{\bf Proof of Theorem~\ref{Theorem: cc multi objective fair kd-tree}}

Multi-objective Fair KD-tree requires a single execution of the ML classifier at the beginning of the algorithm. Therefore, the computational complexity is $\sum_{i\Equal1}^m\mathcal{O}(h_i)$. Once confidence scores are generated, given that $m$ is small, the total required objective computations at every tree level remains $\mathcal{O}(|D|\times \lceil \log(t)\rceil)$ as the combined vector can be calculated in constant time.\\

\end{document}